\documentclass{article}

\usepackage{arxiv}
\usepackage[utf8]{inputenc} 
\usepackage[T1]{fontenc}    
\usepackage{hyperref}       
\usepackage{url}            
\usepackage{booktabs}       
\usepackage{amsfonts}       
\usepackage{amsmath}
\usepackage{amssymb}
\usepackage{nicefrac}       
\usepackage{microtype}      
\usepackage{graphicx}
\usepackage{natbib}
\usepackage{doi}
\usepackage{algorithm}
\usepackage{algpseudocode}

\usepackage{xcolor}
\usepackage{mdframed}

\newif\ifshownew \shownewfalse
\usepackage[finalnew]{trackchanges}

\newmdenv[linecolor=blue!55, linewidth=0.7pt, backgroundcolor=blue!4,
          skipabove=8pt, skipbelow=8pt,
          innerleftmargin=8pt, innerrightmargin=8pt,
          innertopmargin=6pt, innerbottommargin=6pt]{newmaterial}
\long\def\new#1{\ifshownew\begin{newmaterial}#1\end{newmaterial}\else#1\fi}

\title{A memory-based active inference model of DishBrain-like adaptive behaviour}

\date{\today}

\author{
{Aswin Paul\textsuperscript{1}, 
Moein Khajehnejad\textsuperscript{1}, 
Forough Habibollahi\textsuperscript{2}, 
Brett J. Kagan\textsuperscript{2} and 
Adeel Razi\textsuperscript{1,3,4}\thanks{Corresponding author: Adeel Razi: adeel.razi@monash.edu}} \\
\textsuperscript{1} Turner Institute for Brain and Mental Health, School of Psychological Sciences, Monash University, Clayton 3800, Australia
\\
\textsuperscript{2} Cortical Labs Pty Ltd, Melbourne 3056, Australia \\
\textsuperscript{3} Queen Square Institute of Neurology, University College London, London, UK\\
\textsuperscript{4} CIFAR Global Scholars Program, CIFAR, Toronto, Canada \\
}

\renewcommand{\shorttitle}{Memory-based active inference in DishBrain-like agents}

\algnewcommand{\algorithmicand}{\textbf{ and }}
\algnewcommand{\algorithmicor}{\textbf{ or }}
\algnewcommand{\OR}{\algorithmicor}
\algnewcommand{\AND}{\algorithmicand}
\algnewcommand{\var}{\texttt}

\begin{document}
\maketitle

\begin{abstract}
Recent and rapid advances in artificial intelligence (AI) make it increasingly important to understand the foundations of \change{purposeful}{adaptive} behaviour in autonomous agents, especially for building safe and efficient systems. While artificial neural networks have dominated the development of AI, recent work has begun to explore living biological neuronal networks as an alternative substrate for computation. These systems promise remarkable data and sample efficiency and rich dynamics, and may also inspire explainable and biologically plausible models.
Here, we develop an experiment-informed active inference framework to model decision-making in \change{embodied}{closed-loop} agents that mirror experimental setups using biological neurons. \change{Using a generative model grounded in a simulated Pong environment, we systematically study agents that combine memory-based learning and predictive planning.}{Using a generative model whose dimensions are matched to an experiment protocol, we systematically compare three decision-making schemes within this common generative model.} \change{Our simulations demonstrate robust learning in these agents and reveal how memory horizons and planning schemes shape performance and adaptability.}{Under matched episode counts (i.e. total data available for learning) to the in-vitro experiment, our simulations show that agents with short memory horizons reach a level of performance close to that of mouse and human cortical cultures (DishBrain platform), whereas longer memory horizons depart from it substantially. Increasing the planning horizon, by contrast, confers no comparable benefit.} \change{By exposing the dynamics of key model parameters and their relationship to behaviour, this work contributes to the growing field of explainable AI, offering a biologically grounded and scalable approach to understanding purposeful behaviour in artificial and synthetic biological intelligence.}{Because all model parameters are explicit, we can also track the quantities in our generative model that accompany this improvement, such as the risk term and the entropy of the transition and state-action mappings. Together, these results illustrate how active inference offers a formal language for comparing decision-making schemes in similar closed-loop control environments.}
\end{abstract}


\section{Introduction}

The rapid progress of artificial intelligence (AI) has sharpened the need for a principled account of purposeful\add{\footnote{Throughout this paper, we use the term `purposeful' (equivalently, `adaptive') behaviour in a deliberately narrow and operational sense, i.e. a measurable improvement in task performance over the course of a trial, under a generative model, whose structure is fixed in advance.}} behaviour in autonomous agents, particularly for developing systems that are \remove{safe and}controllable. Contemporary AI is largely driven by artificial neural networks (ANNs), which have enabled impressive capabilities but often require large amounts of data and computation, and can be difficult to interpret. In parallel, a growing body of work is turning to biological paradigms, where living neuronal networks (BNNs) offer a complementary route to intelligence, with the promise of high power, data efficiency, and rich adaptive dynamics.

Experimental advancements, such as the DishBrain system \cite{Kagan2022}, place cortical neurons on silicon chips and embody them in a closed-loop Pong environment, demonstrating that even minimal biological substrates can adapt their behaviour in real time through sensory feedback. These studies underpin the emerging field of synthetic biological intelligence (SBI), which treats living neural systems as computational substrates \cite{kagan2023technology}. Alongside related advances in natural computation, such as DNA-based molecular learning systems \cite{Evans2024-dna}, these developments suggest that biologically grounded approaches can inform new models of intelligence that are both efficient and explainable. However, they also raise a central theoretical challenge: how to model, in a principled way, the emergence of purposeful, adaptive behaviour in such systems.

Existing AI frameworks, particularly deep reinforcement learning (RL), have been evaluated against biological cultures in the DishBrain environment. In \cite{khajehnejad2024biologicalneuronscompetedeep, khajehnjad2025}, several deep RL algorithms were compared to DishBrain cultures, revealing that simple neuronal cultures can outperform these algorithms under realistic time \footnote{Algorithmically, we matched the total episode counts across all the agents, which is a proxy for available training data and time.} constraints, especially in terms of sample efficiency. These studies also showed that active inference-based agents can match or surpass the performance of deep RL schemes \add{in data constrained settings}. \change{Nonetheless, deep learning-based models often struggle to capture the adaptive, memory-driven, and embodied nature of biological decision-making, and typically lack the interpretability needed to relate internal computations to observable behaviour.}{Nonetheless, the internal computations of deep learning-based models are not straightforward to relate to observable behaviour, which makes it difficult to ask which mechanism is responsible for a given behavioural course. This motivates models and decision-making schemes whose parameters are explicit and open to interpretations. We note, however, that interpretability of this kind is a property of the model we specify, and is not in itself evidence about the mechanisms used by the technology inspiring such studies.}
 
Active inference, originally developed as a unifying theory of brain function \cite{Friston2010, Friston2012}, offers a promising alternative. It posits that perception, learning, and action arise from the minimisation of (expected) free energy, with behaviour driven by an agent's beliefs about its environment rather than by externally defined reward signals \cite{Friston2009, neftci2019reinforcement}. This provides a probabilistic framework that naturally accommodates uncertainty, \change{embodiment}{closed-loop interaction}, and continual learning. Recent algorithmic developments, such as counterfactual learning (CFL) \cite{Isomura2023, Paul2024} and dynamic programming in expected free energy (DPEFE) \cite{Paul2023, Friston2023inductiveinf}, have further expanded the repertoire of active inference decision-making schemes. Yet, a comprehensive, experiment-informed comparison of these schemes in a biological setting like DishBrain has been lacking.

In this work, we develop an experiment-informed generative model of the DishBrain Pong setup and use it to systematically compare memory-based and planning-based active inference agents. We construct a partially observable Markov decision process (POMDP) that mirrors the key elements of the experimental protocol, including rate- and place-coded stimulation, ball and paddle dynamics, and action constraints. Within this shared generative model, we examine three families of agents: (i) counterfactual learning agents with finite memory horizon ($\text{CFL-}T$), (ii) a classical single-step active inference agent (AIF-1), and (iii) DPEFE agents with variable planning horizon ($\text{DP-}T$). By matching trial durations and episode counts to the in-vitro experiments, we compare in-silico agents directly to mouse and human cortical cultures (MCC and HCC). \add{We emphasise that the contribution of this work is a task-specific modelling study, rather than a new decision-making algorithm or an attempt aimed at deriving new biological hypothesis. Stated precisely, we contribute (i) a generative model whose dimensions are derived from the stimulation protocol of an existing experiment, and (ii) a focused comparison of three active inference decision-making schemes within that shared model, evaluated against in-vitro data under the same episode counts (constrained data).}

Our results show that memory-based CFL agents with short horizons can reproduce and even exceed the performance of biological cultures (under similar data efficient learning constraints), whereas planning-based agents (AIF-1 and DPEFE) exhibit limited gains in this environment. To explain these differences, we analyse key model parameters, including the risk-term and the normalised total entropy (NTE) of probability-encoding parameters such as transition dynamics, state-action mappings, and prior-preference distributions. This analysis reveals distinct signatures of learning and confidence in high-performing agents and highlights limitations of purely planning-based schemes in the Pong task. \change{Together, these findings support a view of DishBrain-like behaviour that emphasises memory-based learning and rapid adaptive restructuring, and illustrate how active inference can provide an interpretable, biologically grounded framework for synthetic biological intelligence and explainable AI.}{Together, these findings show that, in this particular task, a finite state-action update window is sufficient to reproduce the behavioural improvement observed in vitro, whereas deeper planning is not. We emphasise that the aim of this work is not to act a general comparison between memory-based algorithms and planning-based ones, but contrasting their mechanisms under a shared generative model.}

\section{Results}

\begin{table}[ht]
\centering
\small
\caption{Abbreviations used in this paper. Experimental group labels follow the original DishBrain study \cite{Kagan2022}.}
\label{tab:abbrev}
\begin{tabular}{ll@{\hspace{2em}}ll}
\toprule
\multicolumn{2}{c}{\textbf{Models and methods}} & \multicolumn{2}{c}{\textbf{Experimental groups}} \\
\midrule
AI & Artificial intelligence & MCC & Mouse cortical cells \\
ANN & Artificial neural network & HCC & Human cortical cells \\
BNN & Biological neuronal network & CTL & Media-only control \\
SBI & Synthetic biological intelligence & RST & Rest session (spontaneous activity), no feedback \\
RL & Reinforcement learning & IS & In-silico control, random paddle \\
POMDP & Partially observable Markov decision process & & \\
\cmidrule(r){3-4}
EFE & Expected free energy & \multicolumn{2}{c}{\textbf{Measures}} \\
\cmidrule(r){3-4}
KL & Kullback--Leibler (divergence) & TE & Total entropy \\
AIF-1 & Active inference agent, one-step planning & NTE & Normalised total entropy \\
DPEFE & Dynamic programming in expected free energy & RI & Relative improvement \\
DP-$T$ & DPEFE agent with planning horizon $T$ & & \\
CFL-$T$ & Counterfactual learning agent with a & \multicolumn{2}{c}{\textbf{Apparatus}} \\
 & finite state-action update window of length $T$ & & \\
\cmidrule(r){3-4}
 & & HD-MEA & High-density multi-electrode array \\
\bottomrule
\end{tabular}
\end{table}

\subsection{Memory–based counterfactual learning drives strong improvement in performance}
\label{sec:cl_result}

We begin by evaluating active inference agents equipped with the counterfactual learning algorithm \cite{Isomura2023}. We use the memory–augmented variant introduced in \cite{Paul2024}, where the agent learns the $\mathbb{CL}$ mapping from the previous $T$ time steps. We refer to these agents as `CFL-$T$'. Details of the method are provided in Sec.~\ref{section:cl-method}.

To align simulations with the DishBrain setting, each trial is assigned a real-time duration of twenty minutes. Biological trials contained, on average, $69.04 \pm 7.95$ Pong episodes \cite{Kagan2022,khajehnejad2024biologicalneuronscompetedeep}. We therefore simulate $70$ episodes per trial and apply corresponding time labels. For each CFL-$T$ configuration, we run $100$ trials with different random seeds to reduce variability and ensure reproducibility. This enables a direct comparison between in-vitro neuronal cultures and our in-silico agents.

\begin{figure}[htbp]
    \centering
    \includegraphics[width = \textwidth, page=1]{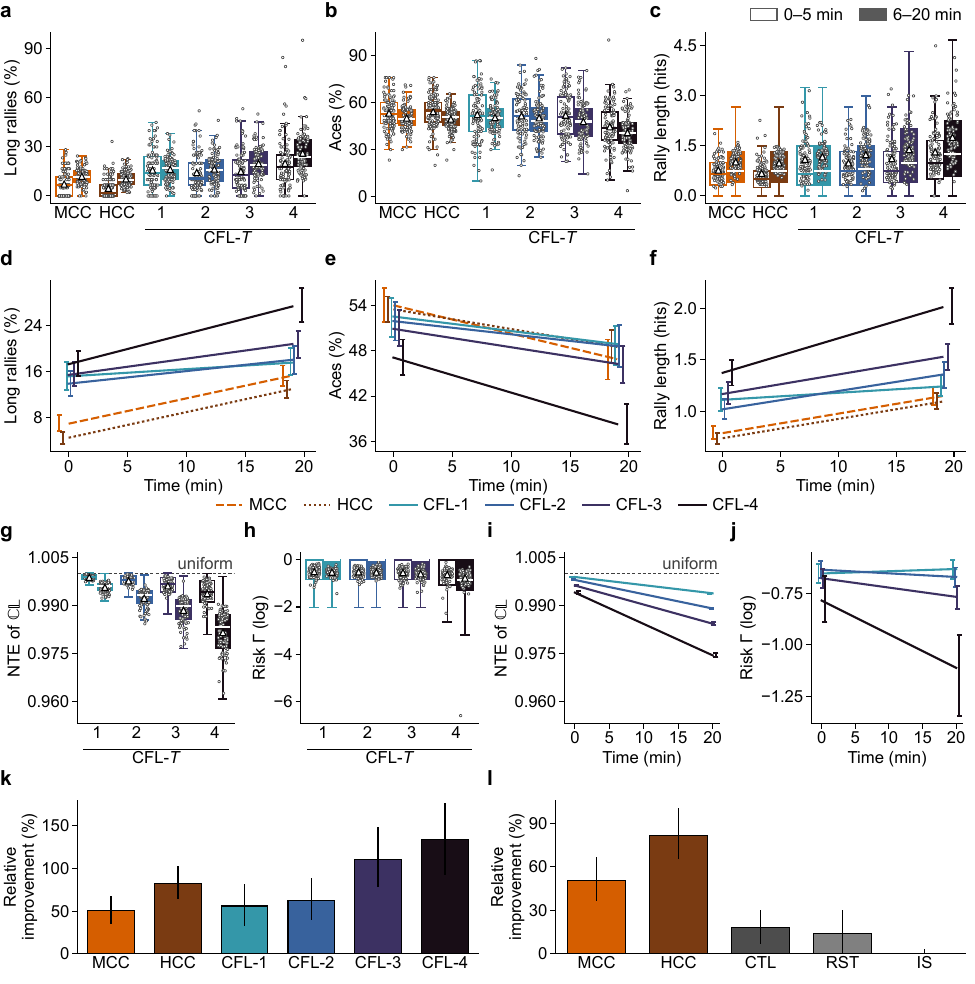}
    \caption{\textbf{Memory horizon influences performance in CFL agents and comparison with MCC and HCC groups.} Relative improvement in key performance metrics for in-silico CFL agents and in-vitro MCC and HCC cultures. Performance is evaluated by comparing the first 5 minutes (open boxes) and last 15 minutes (filled boxes) of each trial. Points on the box plots show individual sessions (in vitro) or trials (in silico); in c, g and h, where the boxes summarise per-minute (c) or per-episode (g, h) values, the points are session or trial means. Metrics include average rally length, percentage of aces (Aces are episodes with score-zero (i.e. immediate miss), lower the better), and percentage of long rallies ($\geq 3$ hits). \textbf{a--c:} CFL-$T$ performance across discrete trial windows. \textbf{d--f:} Linear-regression analysis of continuous-time performance. \textbf{g:} Evolution of the NTE of the $\mathbb{CL}$ vector. \textbf{h:} Evolution of the average risk term $\Gamma_t$. \textbf{i--j:} Continuous-time evolution of NTE and risk. \textbf{k:} Relative improvement of CFL agents across memory horizons. \textbf{l:} Relative improvement of the experimental baseline groups. Group abbreviations: MCC, mouse cortical cells; HCC, human cortical cells; CTL, media-only control; RST, rest session, in which an active culture drove the paddle but received no sensory input; IS, in-silico control, in which the paddle was driven by random noise \mbox{\cite{Kagan2022}}. CFL-$T$ denotes a counterfactual learning agent with a finite state-action update window of length $T$; NTE denotes normalised total entropy. All abbreviations are collected in Table \ref{tab:abbrev}.}
    \label{fig:results-cfl}
\end{figure}

Fig.~\ref{fig:results-cfl} (a--c) shows a consistent performance increase with larger memory horizons. 
\change{Even restricting to low values ($T \leq 4$), CFL agents outperform MCC and HCC in all metrics.}{A more formal analysis supports this in various levels for each agent variants. On the average rally length reached in the second window (a measure of game performance), CFL-2, CFL-3 and CFL-4 all exceed both cultures, and CFL-1 exceeds MCC (Table \mbox{\ref{tab:contrasts}}). On the percentage of aces, however, CFL-1 and CFL-2 are worse than both MCC and HCC (Supplementary Table \mbox{\ref{tab:absolute}}), so the advantage is not uniform across metrics.} 
A memory horizon of $T=4$ already yields substantial gains, although such long-range memory has not been experimentally demonstrated in DishBrain. 
\change{Notably, CFL-1 and CFL-2 perform comparably to MCC and HCC, suggesting that short-term memory alone can account for much of the adaptive behaviour observed in vitro.}{What distinguishes the short horizons is the magnitude of the departure rather than its direction. CFL-1 and CFL-2 exceed the cultures by only $0.13$ to $0.17$ hits, against a culture mean of $0.85$ to $0.88$, whereas CFL-3 exceeds them by roughly $0.56$ and CFL-4 by roughly $1.59$. Short memory horizons therefore stay close to the biological range while longer horizons leave it, which is consistent with short-term memory accounting for much of the adaptive behaviour observed in vitro.} \change{Among all conditions, CFL-3 shows the strongest overall performance across average rally length, long-rally percentage, and aces.}{Among all conditions, CFL-4 shows the strongest overall performance, leading on average rally length, long-rally percentage and aces alike (Supplementary Table \mbox{\ref{tab:absolute}}).} The relative improvements in Fig.~\ref{fig:results-cfl} (k) further show that CFL-4 exhibits the largest gain over time.

\begin{table}[ht]
\centering
\small
\caption{Average rally length reached in the second window (6-20 min) by each in-silico agent, compared with the mouse (MCC) and human (HCC) cortical cultures. Differences and intervals come from a bootstrap that resamples whole replicates; $p$-values are Benjamini-Hochberg corrected across all 18 contrasts.}
\label{tab:contrasts}
\begin{tabular}{llrrrrl}
\toprule
Agent & Culture & $\Delta$ rally & 95\% CI & Hedges' $g$ & $p_{\mathrm{adj}}$ & Outcome \\
\midrule
CFL-1 & MCC & +0.169 & [0.039, 0.336] & +0.33 & 0.013 & agent higher \\
CFL-1 & HCC & +0.139 & [0.000, 0.312] & +0.28 & 0.060 & not distinguishable \\
CFL-2 & MCC & +0.161 & [0.057, 0.268] & +0.45 & 0.005 & agent higher \\
CFL-2 & HCC & +0.131 & [0.017, 0.247] & +0.37 & 0.032 & agent higher \\
CFL-3 & MCC & +0.588 & [0.401, 0.790] & +0.86 & $<$0.001 & agent higher \\
CFL-3 & HCC & +0.558 & [0.366, 0.765] & +0.86 & $<$0.001 & agent higher \\
CFL-4 & MCC & +1.619 & [0.943, 2.737] & +0.47 & $<$0.001 & agent higher \\
CFL-4 & HCC & +1.589 & [0.912, 2.710] & +0.49 & $<$0.001 & agent higher \\
AIF-1 & MCC & -0.049 & [-0.112, 0.003] & -0.26 & 0.076 & not distinguishable \\
AIF-1 & HCC & -0.079 & [-0.155, -0.010] & -0.42 & 0.032 & agent lower \\
DP-2 & MCC & +0.107 & [0.001, 0.226] & +0.28 & 0.060 & not distinguishable \\
DP-2 & HCC & +0.078 & [-0.038, 0.205] & +0.21 & 0.200 & not distinguishable \\
DP-5 & MCC & +0.338 & [0.203, 0.484] & +0.70 & $<$0.001 & agent higher \\
DP-5 & HCC & +0.309 & [0.165, 0.461] & +0.66 & $<$0.001 & agent higher \\
DP-10 & MCC & +0.329 & [0.192, 0.480] & +0.66 & $<$0.001 & agent higher \\
DP-10 & HCC & +0.299 & [0.155, 0.457] & +0.62 & $<$0.001 & agent higher \\
DP-20 & MCC & +0.130 & [0.025, 0.240] & +0.36 & 0.024 & agent higher \\
DP-20 & HCC & +0.100 & [-0.014, 0.219] & +0.28 & 0.090 & not distinguishable \\
\bottomrule
\end{tabular}
\\[4pt]\footnotesize{Positive $\Delta$ favours the agent. ``Not distinguishable'' means the difference is not resolvable at this sample size, and is not evidence that the two groups match; see Supplementary Table \mbox{\ref{tab:contrastsri}} for the corresponding equivalence assessment.}
\end{table}

Together, these findings illustrate that memory-based decision making is highly effective in this task and can exceed the performance of biological cultures. They also motivate closer examination of the interplay between memory and behaviour in DishBrain-like systems. Continuous-time analyses (Fig.~\ref{fig:results-cfl} d--f) reinforce these trends. Section~\ref{sec:explainability} provides an interpretation of parameter dynamics, and Sec.~\ref{sec:add_results} compares CFL agents with other active inference schemes.

\subsection{Planning-based active inference reveals limits compared to memory-driven learning}
\label{sec:add_results}

We next compare CFL agents with two standard families of active inference agents: the classical one-step planner (AIF-1) \cite{sajid2022active, DaCosta2020} and dynamic programming in expected free energy (DPEFE) \cite{Paul2023, Friston2023inductiveinf}. Details appear in Sec.~\ref{section:classical-ai} and Sec.~\ref{section:dpefe}.

\begin{figure}[htbp]
    \centering
    \includegraphics[width = \textwidth, page=3]{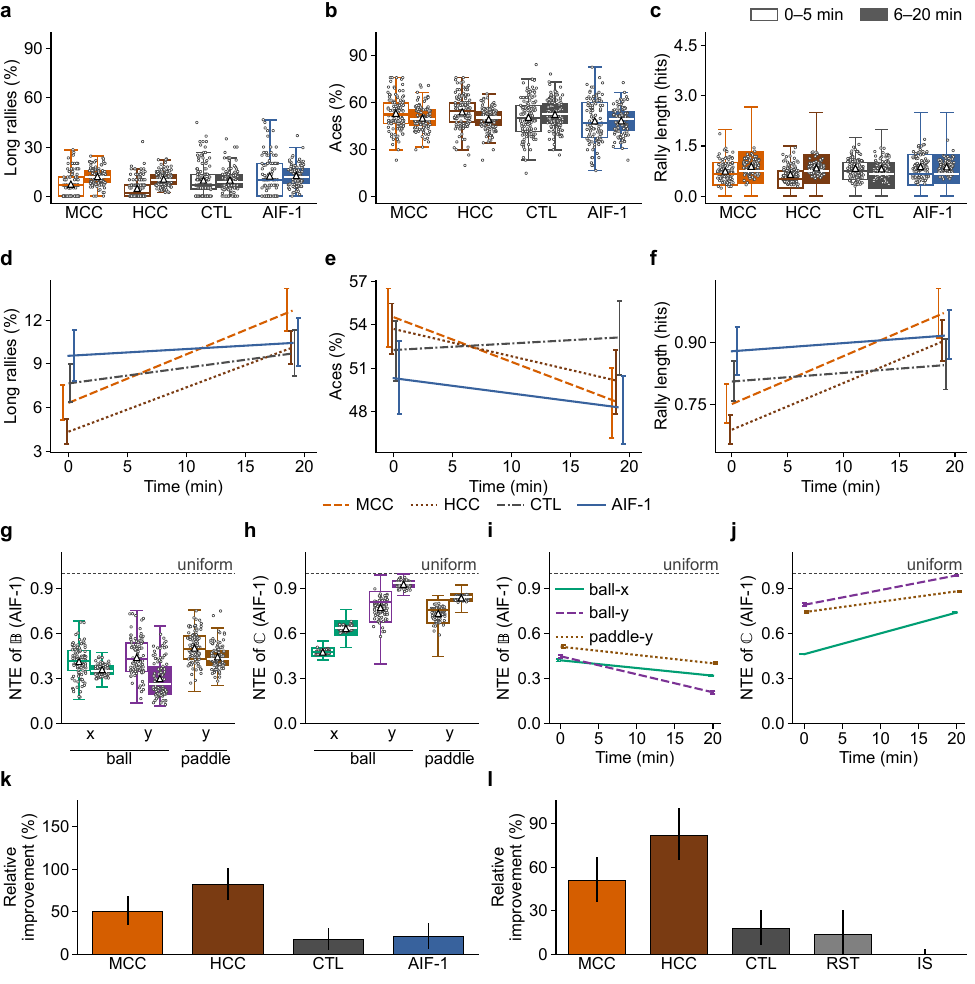}
    \caption{\textbf{Performance of AIF-1 compared to MCC and HCC.} Relative improvement in average rally length, percentage of aces, and percentage of long rallies, evaluated over the first 5 minutes (open boxes) and last 15 minutes (filled boxes). Points on the box plots show individual sessions (in vitro) or trials (in silico); in c, g and h, where the boxes summarise per-minute (c) or per-episode (g, h) values, the points are session or trial means. \textbf{a--c:} Discrete-window performance. \textbf{d--f:} Continuous-time regression. \textbf{g--h:} Evolution of NTE for transition dynamics and prior preferences. \textbf{i--j:} Continuous-time NTE evolution. \textbf{k:} Relative improvement of AIF-1. \textbf{l:} Experimental baselines. AIF-1 denotes the active inference agent with one-step planning; NTE denotes normalised total entropy. Group abbreviations: MCC, mouse cortical cells; HCC, human cortical cells; CTL, media-only control; RST, rest session, in which an active culture drove the paddle but received no sensory input; IS, in-silico control, in which the paddle was driven by random noise \mbox{\cite{Kagan2022}}. See Table \ref{tab:abbrev}.}
    \label{fig:results-aif}
\end{figure}

\begin{figure}[htbp]
    \centering
    \includegraphics[width = \textwidth, page=2]{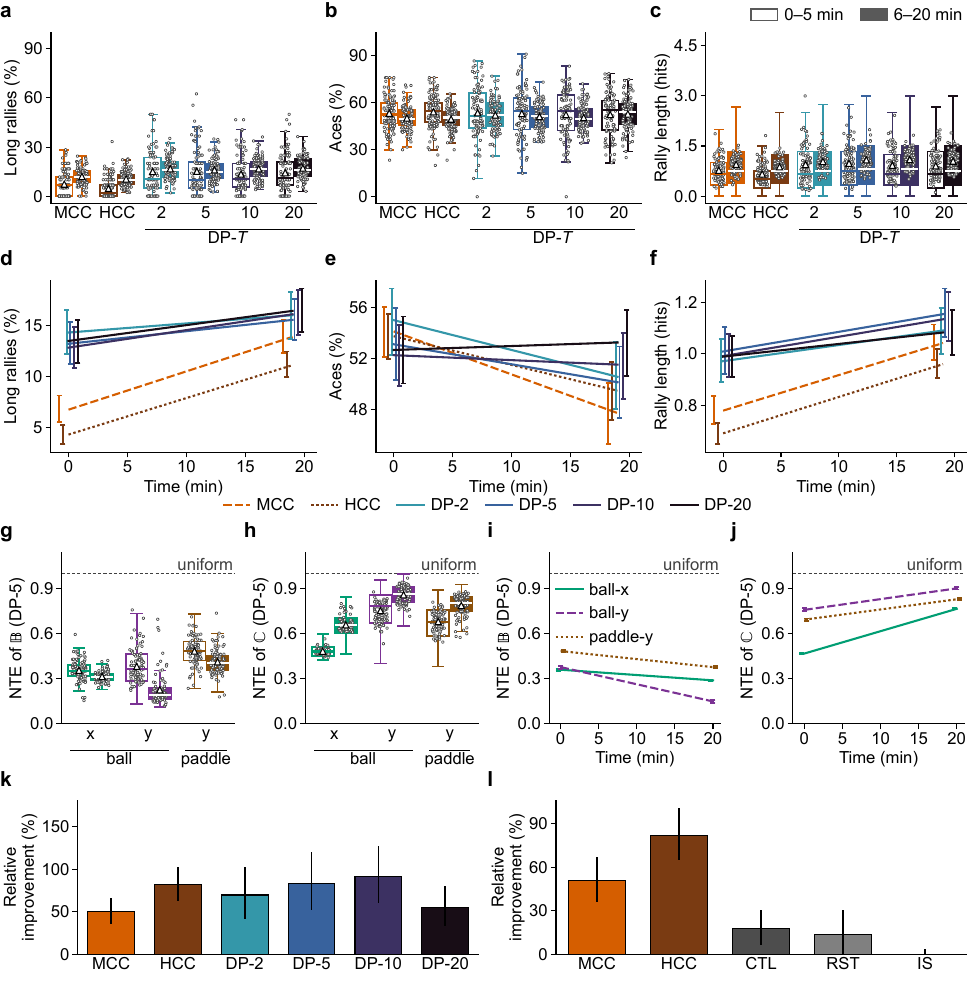}
    \caption{\textbf{Performance of DPEFE agents compared to MCC and HCC.} Structure and metric layout follow Fig.~\ref{fig:results-aif}. Points on the box plots show individual sessions (in vitro) or trials (in silico); in c, g and h, where the boxes summarise per-minute (c) or per-episode (g, h) values, the points are session or trial means. \textbf{a--c:} DP-$T$ performance. \textbf{d--f:} Continuous-time regression. \textbf{g--h:} NTE evolution for transition and preference distributions. \textbf{i--j:} Continuous-time NTE evolution. \textbf{k:} Relative improvement of DP-$T$ agents. \textbf{l:} Experimental baselines. DP-$T$ denotes a dynamic-programming expected-free-energy agent with planning horizon $T$; NTE denotes normalised total entropy. Group abbreviations: MCC, mouse cortical cells; HCC, human cortical cells; CTL, media-only control; RST, rest session, in which an active culture drove the paddle but received no sensory input; IS, in-silico control, in which the paddle was driven by random noise \mbox{\cite{Kagan2022}}. See Table \ref{tab:abbrev}.}
    \label{fig:results-dpefe}
\end{figure}

The comparative results are summarised in Fig.~\ref{fig:results-aif} and Fig.~\ref{fig:results-dpefe}. We select agent configurations whose performance lies within the range of MCC and HCC to enable meaningful comparison. Three main observations follow:

\begin{itemize}
    \item \textbf{DPEFE agents:}  
    Fig.~\ref{fig:results-dpefe} (a--f) shows that DP-$T$ agents improve over the course of a trial. On absolute rally length, DP-5 and DP-10 exceed both cultures, whereas DP-2 is not distinguishable from either (Table \ref{tab:contrasts}); planning is therefore not ineffective in this task. 
    \remove{Increasing the planning horizon does not yield substantial gains, and performance declines for horizons above $T=10$, suggesting an over-planning effect.}\new{Increasing the planning horizon does not yield substantial gains. Regressing relative improvement on $\log T$ gives a slope of $-3.1$ percentage points that is not distinguishable from zero ($95\%$ CI $[-16.7, 10.4]$, $p = 0.65$), and the replicate-level intervals for DP-10 $[47.1, 88.3]$ and DP-20 $[23.6, 57.5]$ overlap; the corresponding point estimates are given in Supplementary Table~\ref{tab:level}. The data therefore support an absence of benefit from deeper planning, rather than an active cost to it.} 
    \add{The reason is not environment variability, since it is deterministic. But that the agent plans through its estimate of the environment rather than through the environment itself. Composing the fitted transition parameters with themselves spreads whatever probability mass is misplaced at one step, so the normalised entropy of the $h$-step-ahead prediction rises from $0.32$ at $h = 1$ to $0.71$ at $h = 5$ and $0.87$ at $h = 20$ (Supplementary Table \mbox{\ref{tab:planningdepth}}; Sec.~\mbox{\ref{sec:predictive-depth}}). A rollout deep enough to reach the horizons of DP-10 and DP-20 is therefore scoring policies against a prediction that is close to uninformative, which is why the additional depth returns nothing.} 
    The relative improvements in Fig.~\ref{fig:results-dpefe} (k) confirm that planning depth alone is insufficient for meaningful improvement in this task.
    
    \item \textbf{AIF-1 agents:}  
    Fig.~\ref{fig:results-aif} shows that AIF-1 achieves only small gains over time, comparable to the experimental control condition. 
    \add{The mixed-effects analysis is in fact stronger than this: AIF-1 shows no detectable change in average rally length across the trial ($-0.016$, $p_{\mathrm{adj}} = 0.71$; Supplementary Table \mbox{\ref{tab:within}}), and it falls significantly below HCC on absolute rally length ($p_{\mathrm{adj}} = 0.032$).} With one-step planning, the agent fails to acquire meaningful task improvements. 
    \add{Setting the prior preferences and the initial state prior to uniform, as in the other two schemes, leaves second-window rally length unchanged (Sec.~\mbox{\ref{sec:priorcheck}}). Hence this is not a consequence of the particular prior preference defined.}

    \item \textbf{Memory vs planning:}  
    \change{Together, these results indicate that memory-based strategies are more effective for real-time interactive environments like Pong than planning-focused algorithms. This supports the hypothesis that DishBrain-like behaviour is better captured by short-horizon memory mechanisms than by deep future-state evaluation.}{Together, these results indicate that the advantage of the memory-based scheme tracks the memory horizon, rather than the scheme itself. Only the longer-horizon agents CFL-3 and CFL-4 exceed the planning-based agents. We note that this is a statement about the agents we simulate, and not about the mechanism used by the neuronal cultures.}
\end{itemize}

\subsection{Baselines outside active inference and observations}
\label{sec:baselines}

\new{
The comparisons above are between active inference schemes. We add four baselines, each run through the same environment loop, reset protocol and window-splitting rule as the agents above. The baselines are a uniformly random controller, a reactive heuristic that moves the paddle towards the ball (by construction), tabular Dyna-Q over the same $2432$ states, and Dyna-Q with ten replay steps per transition. Implementation details are given in Sec.~\ref{sec:baseline-methods}, and the results in Supplementary Tables \ref{tab:baselines} and \ref{tab:baseline-contrasts}. The observations from the baseline agents are noted below.

First, the random controller does not improve across the trial ($+0.02$ hits, $95\%$ CI $[-0.04, 0.08]$, $p_{\mathrm{adj}} = 0.640$) as expected. This is the calibration our improvement metric requires. A controller that cannot learn shows no gain under the window rule we use, so the improvements reported for the cultures and for the learning agents are not an artefact of how the two windows are defined. It is also the in-silico control of the original experiment, in which the paddle was driven by random noise \cite{Kagan2022}, so the floor against which we read the other agents is the one that study itself adopted rather than a construction of ours.

Second, the reactive heuristic reaches an average rally length of $12.07$ $[11.28, 12.89]$, against $2.47$ for CFL-4 and $0.85$ to $0.88$ for the two cultures (Hedges' $g = 2.10$ against CFL-4). Pong under full observability is therefore solvable by a controller that does not learn at all. Performing well at this task in itself need not be evidence of sophisticated computation, hence there is a clear need to differentiate learning from performance.

Third, tabular Dyna-Q is indistinguishable from CFL-1 ($p_{\mathrm{adj}} = 0.108$), exceeds CFL-2 ($p_{\mathrm{adj}} = 0.018$), is indistinguishable from DP-5 ($p_{\mathrm{adj}} = 0.861$), and improves within a trial much as the cultures do ($+0.28$ $[0.20, 0.37]$). Only CFL-3 and CFL-4 exceed it. The advantage we report at short memory horizons is therefore not specific to active inference, and within-trial improvement alone does not discriminate between accounts of the biological data. Taken with the observations in Sec.~\ref{sec:add_results}, these comparisons show that performance improves with the increasing CFL update horizon in this task, with CFL exceeding the non-active-inference baseline only once $T \geq 3$.

One further observation concerns AIF-1. It is indistinguishable from the random controller on absolute rally length ($0.80$ against $0.80$, $p_{\mathrm{adj}} = 0.952$) and shows no improvement across the trial ($p_{\mathrm{adj}} = 0.682$). One-step planning does not merely underperform the other schemes in this task; it does not leave the floor set by random action.
}

\subsection{Experimental noise and partial observability}
\label{sec:noise}

\new{
To understand the effects of noise, we repeated the simulations with the likelihood mixed towards uniform, and separately with the executed action resampled at random, as described in Sec.~\ref{sec:noise-methods}. Results are given in Supplementary Table \ref{tab:noise}. We note our observations below.

Observation noise does not degrade the memory-based agent. It improves it in some cases. CFL-4 reaches an average rally length of $2.39$ in the control condition, against $6.29$ and $4.75$ under mild and strong observation noise respectively. This is a shift of the whole distribution rather than of its tail: the median rises from $1.83$ to $2.81$ and $3.27$, and $73\%$ of seeds exceed the control median. The improvement is not accompanied by a more informed state-action mapping, since the normalised total entropy of $\mathbb{CL}$ falls by between $4.4$ and $4.6$ per cent of its maximum in every condition, the control included. We are able to state, then, that the result of Sec.~\ref{sec:cl_result} is not an artefact of full observability.

For the planning agent the picture differs, and the evidence is weaker. DP-5 reaches $1.30$ in the control condition against $1.12$ and $0.90$ under the two levels of observation noise, and none of these differences is distinguishable from zero. We report them as indicative and draw no conclusion from them.

Unreliable action execution is the variation that degrades performance, and it does so for both schemes. CFL-4 falls from $2.39$ to $1.46$, and DP-5 from $1.30$ to $0.73$, its within-trial change becoming negative ($-0.49$, $p = 0.018$). This is the one condition in which the planning agent is worse at the end of a trial than at the beginning, and it is significant despite the small number of seeds. Corrupting the link between a chosen action and its execution therefore costs both schemes their improvement, whereas corrupting the observation costs only the planning scheme, if it costs anything at all.
}

\subsection{Model parameters reveal explainable differences across memory and planning-based agents}
\label{sec:explainability}

Because all model parameters are explicitly represented, we can analyse how they evolve during learning. We use the same temporal framing as the biological experiment.

For CFL agents, Fig.~\ref{fig:results-cfl} (h) shows the evolution of the average risk term $\Gamma_t$ (log scale). A marked drop occurs only for CFL-4, consistent with its superior performance in Fig.~\ref{fig:results-cfl} (a--f). To analyse probability-encoding parameters such as $\mathbb{B}$, $\mathbb{CL}$, and $\mathbb{C}$, we introduce the Normalised Total Entropy (NTE) metric (Sec.~\ref{sec:nte}). Lower NTE indicates a more informed distribution relative to an uninformed prior. Fig.~\ref{fig:results-cfl} (g--j) shows that NTE for the $\mathbb{CL}$ vector 
\change{decreases most prominently for higher memory horizons, supporting the emergence of goal-directed learning.}{decreases monotonically with the memory horizon.} 
\add{Measured against the analytic maximum, however, the change is small: the mapping falls by $1.2\%$ for CFL-1 and by $4.3\%$ for CFL-4, and so remains close to uninformed throughout (Supplementary Table \mbox{\ref{tab:nte}}). What the horizon orders is the size of a small departure, not a transition to a determined mapping.}

For DPEFE agents, Fig.~\ref{fig:results-dpefe} (g--j) shows decreasing NTE for transition dynamics $\mathbb{B}$, indicating improved environmental understanding. However, NTE for prior preferences $\mathbb{C}$ increases, which is consistent with the structure of the Pong task: no specific paddle or ball position is inherently preferred, and only successful defence is rewarded. The agent therefore cannot form stable preferences over particular observations.

The same pattern appears for AIF-1 (Fig.~\ref{fig:results-aif} g--j). Transition dynamics become more informed, but prior preferences become less structured over time.

\new{
Two observations follow from measuring all three schemes against a common reference. The first is that the planning schemes move their parameters far more than the memory-based scheme does: $\mathbb{B}$ falls by $68\%$ to $71\%$ of its maximum for the DP-$T$ agents, against the $1.2\%$ to $4.3\%$ of the $\mathbb{CL}$ mapping. The second is that the size of that movement does not order the agents by performance. AIF-1 reduces the entropy of $\mathbb{B}$ by $83.8\%$, the largest reduction of any agent, while reaching an average rally length indistinguishable from a random controller (Sec.~\ref{sec:baselines}). Among the planning agents the largest reduction belongs to DP-2, which is the weakest of them. We tested directly whether seeds whose parameters became more informed were the seeds that improved, and found no association that survives correction for any agent (Supplementary Table \ref{tab:ntepredict}).
}

A full exploration of parameter dynamics and alternative generative model structures is left to future work. \change{The analyses demonstrated here illustrate how mechanistic interpretability in active inference can yield insight into the basis of adaptive performance in both artificial and biological agents.}{What the analyses demonstrated here establish is more focussed than an account of performance.} 
\add{Because every parameter is explicitly represented, the internal state of the agent is available for inspection at any point in a trial, and quantities such as NTE can be computed on it directly. The dissociation described above is itself a product of that access, in that it is a question one can only ask of a model whose parameters are open to measurement. Interpretability of this kind supports deeper analysis of a decision-making scheme, rather than, on its own, explaining the behaviour that a scheme produces every time.}

\subsection{Discriminative predictions and further observations}
\new{
Because the three schemes differ in what they require of the agent, they make different predictions about how behaviour should degrade under manipulation. These predictions are testable in the existing DishBrain preparation without new modelling. We set out four, ordered from the least to the most demanding experimentally.

First, if behaviour depends on a finite window of recent state-action pairs, then corrupting that window, for instance by interleaving stimulation unrelated to the current game state, should degrade performance in proportion to the fraction of the window disrupted, and performance should recover over a comparable number of steps. A scheme relying on a learned transition model should be less affected, since $\mathbb{B}$ accumulates over a much longer timescale than the window does. 

Second, temporally shuffling sensory feedback within a session should penalise the counterfactual scheme disproportionately, because the association it forms is between the state immediately preceding an action and that action, whereas a scheme scoring rollouts against a preference distribution is comparatively insensitive to local ordering. 

Third, removing paddle-position feedback should collapse the counterfactual scheme towards chance more sharply than it collapses a planning scheme, which can partially recover paddle position from its transition dynamics.

The fourth prediction is the most discriminative, and follows directly from the trade-off set out in Sec.~\ref{section:pomdp}. Introducing ball trajectories not previously encountered should favour the planning schemes, which can evaluate unvisited states through $\mathbb{B}$, over the counterfactual scheme, which can only reproduce regularities already present in its mapping. A culture that generalises to novel trajectories would therefore be less consistent with a purely history-based account, whereas a selective loss of performance under such trajectories would be more consistent with one. Neither outcome, however, would uniquely identify the underlying mechanism.

We also note what would be required to connect the model variables of Sec.~\ref{sec:explainability} to measurable neural quantities, since the present study cannot do so. It would need a session-level pairing between fitted model parameters and recorded activity from the same session. Our agents and the cultures are not paired in this way, as the agents are indexed by random seed and the cultures by chip, so no such correspondence can be estimated from these data however they are analysed. Establishing it would require fitting the generative model to individual culture sessions rather than simulating it alongside them.

Next, the behaviour of the prior preferences deserves comment, because it bears on how far the comparison between schemes can be taken. In both planning schemes the entropy of $\mathbb{C}$ rises over the course of a trial rather than falling, and we read this as a property of how preferences are specified in our model rather than as a failure of planning in general. Success at Pong is a relation between two quantities, the position of the ball and the position of the paddle, whereas $\mathbb{C}$ is specified here as three independent distributions, one over each observation modality. A parametrisation of that form can express a preference for particular ball or paddle positions taken separately, but it cannot express a preference for the paddle being aligned with the ball, which is the quantity that actually predicts a hit. There is accordingly no configuration of absolute positions on which the agent could usefully settle. Encoding a relational preference would require a modality defined over a derived quantity, such as the ball-paddle offset, which changes the structure of the generative model rather than the values of its parameters, and we therefore leave it outside the present comparison.

This bears on how our planning results should be read. The horizon at which the DP-$T$ agents cease to improve follows from the decay of predictive information in the learned transition parameters (Sec.~\ref{sec:predictive-depth}); the weak performance of AIF-1 is not attributable to its prior specification (Sec.~\ref{sec:priorcheck}); and the preference structure just described bounds what any of the planning agents can represent. These are three separate constraints, and our data do not allow us to apportion the shortfall between them. We therefore report that the planning schemes underperform in this task and under this model, and we decline the more general conclusion that planning is less suited than memory to environments of this kind.
}

\section{Discussion}
\label{sec:discussions}

We have shown that an experiment-informed active inference framework 
\change{can effectively model}{can be used to model} purposeful decision-making in a synthetic biological intelligence context. Using a generative model aligned with the DishBrain Pong environment, we demonstrated that agents combining active inference with memory-based counterfactual learning achieve robust learning and adaptation in real time. 
\change{Within this setting, active inference offers a biologically plausible and inherently interpretable account of how systems resembling biological neuronal networks can sustain continuous learning, real-time responsiveness, and purposeful behaviour. By unifying perception, action, and learning under a single probabilistic model, it emerges as a strong alternative to traditional machine learning paradigms for developing biologically inspired AI.}{Within this setting, active inference provides a formal and inspectable account of how an agent can sustain continuous learning and real-time responsiveness while remaining coupled to its environment. We note that this is an account of the model we specify, and not of the biological substrate. Our results concern the behaviour of these agents in a single task.}

\change{These advantages are particularly evident when considered alongside}{These considerations sit alongside} the remarkable sample efficiency of biological systems, which many artificial systems, including deep reinforcement learning (RL) agents, still struggle to match \cite{khajehnjad2025}. Our adoption of active inference is motivated not only by a desire to capture this efficiency, but also by the need for biological plausibility and interpretability
\change{, domains where opaque and heavily tuned RL models often fall short}{, which are harder to obtain from models with a large number of tuned parameters} \cite{Friston2010, Friston2012}. In systems like DishBrain, emergent behaviour arises from the interplay between external stimuli and living neuronal dynamics. A probabilistic framework that explicitly encodes uncertainty and preferences, as active inference does, is therefore well suited to modelling how such systems learn to act purposefully in uncertain environments.

Our analysis of key model parameters provides a \change{mechanistic window into these learning dynamics}{window into the learning dynamics of the model}. The evolution of the risk term ($\Gamma$) and the normalised total entropy (NTE) of the $\mathbb{CL}$ vector reveals how high-performing CFL agents progressively reduce uncertainty and refine their policies \cite{Paul2024}. In particular, the observed reductions in risk and NTE for agents with higher memory horizons reflect increasing confidence and more informed decision-making over time, consistent with the emergence of goal-directed behaviour. \change{Because these quantities are explicitly accessible in the model, they support the development of safe and transparent AI systems, where one can follow how uncertainty is minimised through adaptive exploration, rather than treating performance as a black-box outcome. In this way, the framework helps bridge experimental neuroscience with formal theories of intelligence, grounding behavioural improvements in interpretable parameter dynamics.}{Because these quantities are explicitly accessible in the model, one can follow how uncertainty is minimised over the course of a trial, rather than treating performance as a black-box outcome~\mbox{\cite{albarracin2023-xaif}}. We emphasise that this is formal interpretability of a generative model that we have specified ourselves, and it should not be conflated with a mechanistic explanation of the biological substrate. Whether these model variables correspond to measurable neural quantities in the cultures is a separate question, and one that the present analysis does not settle.}

At the same time, our findings highlight important open questions about the biological instantiation of memory mechanisms. In our simulations, CFL agents with extended memory horizons show enhanced performance, but current experimental evidence from DishBrain points to rapid network reorganisation and population-wide dynamics rather than explicit long-term memory recall \cite{Kagan2022, habibollahi2023critical}. This tension suggests a key direction for future work: clarifying whether sophisticated memory, as modelled here, is necessary for biological learning, or whether rapid adaptive restructuring alone can underpin the observed behaviours. Distinguishing between these possibilities will require a closer integration of modelling and experiment, for example by probing how different stimulation protocols or perturbations affect effective memory and performance in neuronal cultures.

It is worth being precise here about what the counterfactual update would and would not require of a biological substrate, since the buffer is an implementation device and not itself the hypothesis. The update in Eq.\ref{cp-clmodeleqn} adds an outer product of an action with a state estimate, scaled by a single scalar risk signal. In that form it is an associative, Hebbian-like strengthening of a state-action pairing, gated by a global gain, and is conceptually reminiscent of neuromodulated three-factor plasticity rules used to model reward-modulated learning in neural circuits \cite{Friston2009}. What the mapping encodes is therefore distributed across the state-action association itself, and realising it does not require a stored list of past instances. It requires only that recent activity leave a trace on which plasticity can act. The finite update window in our implementation is one way of expressing that requirement in an algorithm; it is not a claim that cultures maintain an explicit buffer.

Read this way, the apparent tension between a memory buffer and network reorganisation is less sharp than it first appears. Evidence from DishBrain and related preparations points to changes in population-level connectivity and critical dynamics rather than to explicit recall \cite{Kagan2022, habibollahi2023critical, khajehnejad2026graph}, and that evidence is compatible with an associative-plasticity reading of the $\mathbb{CL}$ mapping while being difficult to reconcile with a literal buffer. We would add one caveat on timescale: over the twenty minutes of a single session, neurogenesis is not a plausible substrate, and the relevant candidates are synaptic and network-level plasticity operating on the seconds-to-minutes scale over which the cultures are observed to improve.

\add{
A question that naturally follows is whether schemes of this kind are feasible for tasks more demanding than Pong. The computational costs are known, and are properties of the algorithms rather than results of ours. Enumerating policies makes the classical formulation exponential in the planning horizon, which is why a one-step agent is the only classical variant we can run at this state-space size. Evaluating expected free energy recursively backwards in time is linear in the horizon, and the counterfactual scheme dispenses with forward evaluation altogether. What our data cannot support is any claim about performance in more demanding environments. Pong under full observability is solved better by a reactive controller than by any agent we simulate, and a generic tabular reinforcement learner matches the shorter memory horizons, so this task does not discriminate between the schemes in the way a more complex one might. A demonstration of feasibility would require an environment in which the structure the planning schemes represent struggle.}

Future research should therefore focus on disentangling the respective roles of memory, planning, and fast adaptive dynamics in both synthetic and biological agents. A deeper understanding of these mechanisms will be crucial for refining generative models and for designing hybrid systems that leverage the adaptive strengths of biological intelligence while maintaining the explanatory power of computational frameworks such as active inference. 
\change{Ultimately, the present work contributes to a more integrated understanding of purposeful behaviour, supporting the development of AI that is not only capable and sample-efficient, but also transparent and firmly grounded in biological principles.}{Ultimately, the present work is a focused modelling study, and its contribution is to make the comparison between these decision-making schemes explicit and reproducible within a generative model that is tied to an existing biological preparation.}

\section{Methods}
\label{sec:methods}

\subsection{Generative models based on POMDPs}
\label{section:pomdp}

A generative model is a scaled-down version of the natural world, but not too simple \cite{FRISTON20231}. The purpose is to enable the agent to have prior expectations (about incoming observations) and simultaneously predict future observations to control them in its favour. So, active inference, at its core, is a model-based framework. However, it does not impose a particular structure on such models. Several models are viable for these problems, including Partially Observable Markov Decision Processes (POMDPs) \cite{Kaelbling1998, Lovejoy1991} and Gaussian mixture models \cite{gmm_ieee2005}, depending on the specific requirements of the task. Generative models are gaining popularity in machine learning for various reasons, like the emergence of ChatGPT \cite{chat_gpt_2023}. Their popularity is also emerging in fields including weather modelling \cite{gm_weather_2023} and protein engineering \cite{Ingraham2023}. In this paper, we stick to the POMDP architecture, assuming the agent encodes a discrete world model \cite{DaCosta2020} (See Sec.\ref{section:pomdp}).

POMDPs offer a universal structure to model discrete state-space environments where parameters can be expressed as tractable categorical distributions \citep{Kaelbling1998}. A POMDP can be formally defined as a tuple of finite sets $(S, O, U, \mathbb{B},\mathbb{A})$:

\begin{itemize}
    \item [$\circ$] $s \in S:$ $S$ is a set of hidden states ($s$) causing observations $o$.
    \item [$\circ$]$o \in O:$ $O$ is a set of observations, where $o=s$, in the fully observable setting. In a partially observable setting,  $o=f(s)$.
    \item [$\circ$] $u \in U:$ $U$ is a set of actions ($u$).
    \item [$\circ$]$\mathbb{B}:$ encodes the one-step transition dynamics, $P(s_{t} \vert s_{t-1}, u_{t-1})$ i.e., the probability of transitioning to state $s_{t}$ at time $t$ given that action $u_{t-1}$ is taken in state $s_{t-1}$ at time $t-1$.
    \item [$\circ$] $\mathbb{A}:$ encodes the likelihood mapping, $P(o_{\tau} \vert s_{\tau})$, for  the partially observable setting.
    \item [$\circ$] $\mathbb{D}:$ Encodes the prior of the agent about the hidden state factor $s$.
    \item [$\circ$] $\mathbb{E}:$ Encodes the prior of the agent about actions $u$.
    
\end{itemize}

In a POMDP, the hidden states ($s$) generate observations ($o$) through the likelihood mapping ($\mathbb{A}$) in the form of a categorical distribution, $P(o_{\tau} \vert s_{\tau}) = \mathrm{Cat}(\mathbb{A} \times s_\tau)$.

$\mathbb{B}$ is a collection of square matrices $\mathbb{B}_{u}$, where $\mathbb{B}_{u}$ represents transition dynamics $P(s_{t} \vert s_{t-1}, u_{t-1} = u$): ($\mathbb{B}$) determines the dynamics of $s$  given the agent's action $u$ as $P(s_{t} \vert s_{t-1}, u_{t-1}) = \mathrm{Cat}(\mathbb{B}_{u_{t-1}} \times s_{t-1})$. In $\left[\mathbb{A} \times s_\tau \right]$ and $\left[ \mathbb{B}_{u_\tau} \times s_\tau \right]$, $s_\tau$ is represented as a one-hot vector that is multiplied through regular matrix multiplication \footnote{One-hot is a group of bits among which the legal combinations of values are only those with a single high (1) bit and all the others low (0). Here, the bit (1) is allocated to the state $s= s_\tau$ }.

The \textit{Markovianity} of POMDPs means that state transitions are independent of history (i.e. state $s_{t}$ only depends upon the state-action pair $(s_{t-1}, u_{t-1})$ and not $s_{t-2}, ~u_{t-2}$ etc.). The generative model can be summarised as follows,
\begin{equation}
P(o_{1:t},s_{1:t},u_{1:t}) = P(\mathbb{A}) P(\mathbb{B}) P(\mathbb{D}) P(\mathbb{E}) \prod_{\tau=1}^t P(o_{\tau} \vert s_{\tau}, \mathbb{A}) \prod_{\tau=2}^t P(s_{\tau} \vert s_{\tau-1}, u_{\tau-1},\mathbb{B}).
\end{equation}

So, from the agent's perspective, when encountering a stream of observations in time, such as $(o_{1}, o_{2}, o_{3}, ..., o_{t})$, as a consequence of performing a series of actions $(u_{1}, u_{2}, u_{3}, ..., u_{t-1})$, the generative model quantitatively couples and quantifies the causal relationship from action to observation through some assumed hidden states of the environment. These are called `hidden' states because the agent cannot directly observe the hidden states $s$ in the environment. The agent maintains beliefs about $s$, using the observations $o$. In the experiment (\cite{Kagan2022}), the observations directly represent the state parameters, such as the ball position without noise. Hence, the problem is more control-oriented than a perception problem. So, throughout the paper, we assume $\mathbb{A}:$ that encodes the likelihood mapping ($P(o_{\tau} \vert s_{\tau})$) is an identity distribution.

Based on this representation, an agent can now attempt to optimise its actions to keep receiving preferred observations. So far, the introduced generative model has no concept of `preference' and `goal'. In the next few sections, we focus on decision-making schemes in active inference, addressing variations of algorithms used to model `goal-directed' (i.e. purposeful) behaviour. In the next section, we lay out the experiment-informed generative model we designed and used throughout the paper.

\subsection{Experiment informed generative model}
\label{section:exp_genmodel}

\begin{figure}[ht]
    \centering
    \includegraphics[width = \textwidth]{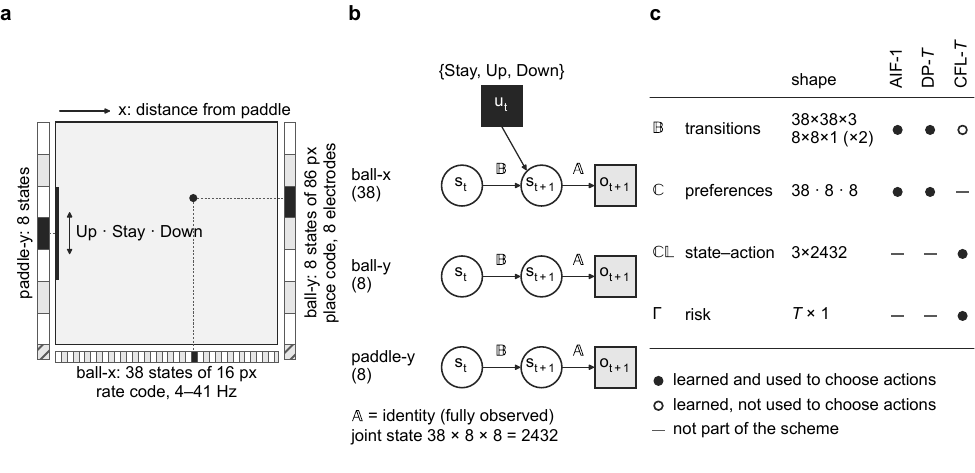}
    \caption{\textbf{Experiment-informed generative model shared by all agents.} \textbf{a:} The Pong court to scale ($600 \times 600$~px, paddle $250$~px). Each state factor discretises one game variable: `ball-x' into 38 bins of 16~px, communicated to DishBrain as a rate code ($4$--$41$~Hz); `ball-y' into 8 bins of 86~px, place-coded over 8 stimulation electrodes; and `paddle-y', the paddle centre, into the same 8 bins. Filled cells mark the bins of the example ball and paddle; hatched cells lie beyond the court edge. \textbf{b:} The generative model over one time step. Each factor has its own transition model $\mathbb{B}$ and an identity likelihood $\mathbb{A}$ (fully observed states); the action $u_t \in \{\mathrm{Stay}, \mathrm{Up}, \mathrm{Down}\}$ enters the `ball-x' transition. \textbf{c:} Parameters of interest, their array shapes, and the schemes that learn and use them. All abbreviations are collected in Table \ref{tab:abbrev}.}
    \label{fig:gen_model}
\end{figure}

\paragraph{State-outcome modalities}
The information about the Pong game states was communicated to the population of neurons via the DishBrain system using biologically safe voltage and frequency levels in the experiment. We assume three state-observation modalities corresponding to the input signals in the generative model: ' ball-x', `ball-y', and `paddle-y'. They represent the ball's $x$ and $y$ coordinates and the $y$ coordinate of the middle of the paddle, respectively. In the experiment, the $x$ coordinate of the ball represents its distance from the paddle along the x-axis. The experiment communicates it through a rate-coded component of 38 distinct frequencies ranging from $4-41$ Hertz. The $y$ coordinate of the ball is transmitted through one of the 8 evenly distributed stimulation electrodes on the y-axis of HD-MEA (See Fig. \ref{fig:gen_model} a). Using this information, we can set up the structure of the generative model.

\paragraph{Structure}

We assume the state space of the generative model has $2432$ states (See Fig. \ref{fig:gen_model} b) in correspondence with the dimensions mentioned above. $38$ distinct states representing `ball-x', and $8$ different states for each of the `ball-y' and `paddle-y', respectively. 
The action space has three elements, namely `Up', `Down', and `Stay', the same as in the simulated game environment of DishBrain. Finally, the observations are assumed to be fully observable states, as there is no noise in communicating the states to the chip.

\paragraph{Parameters of interest}
In this work, we are particularly interested in some parameters of the POMDP critical to decision-making schemes. For the full structure of the generative model, refer to Sec.\ref{section:pomdp} in Methods. The parameters of interest are: 
\begin{itemize}
    \item $\mathbb{B}$: The transition dynamics represent the knowledge about state transitions in the system. An informed $\mathbb{B}$ helps predict the future states in the hope of controlling them.
    \item $\mathbb{CL}$: The counterfactual learning method \cite{Isomura2023, Paul2024} (CFL) which depends on the state-action mapping $\mathbb{CL}$ for decision-making.
    \item $\Gamma$: The Risk term in counterfactual learning represents the agent's confidence. A decrease/increase in $\Gamma$ can indicate a change in strategy, influencing the learning of $\mathbb{CL}$ mapping used in decision-making.
    \item $\mathbb{C}$: The prior preference distribution which drives control in the classical active inference agents.
\end{itemize}

We limited ourselves to the discrete-time, discrete-state POMDP architecture captivated by their explicit structure. Several candidates exist, such as Gaussian mixture models \cite{friston2018deep} for continuous problems and ANNs \cite{Millidge2020a} for representing the dynamics for high dimensional problems. An interesting observation made while experimenting with active inference models is the influence of tiny changes to the structure and dimensions of the model in performance. The freedom of model structure comes with the cost of fine-tuning details but opens up endless ground for exploration. Algorithms for self-learning the structure of generative models are a promising direction to pursue \cite{friston2023supervised}.

\subsection{Decision making schemes chosen in the work}
\new{
The generative model set out above is shared by every in-silico agent in this paper. What separates the three decision-making schemes is not what an agent believes about the world, but how it converts those beliefs into an action. The next three subsections present them in order of decreasing reliance on explicit forward planning, and it is worth stating the theoretical trade-off between them before any results are considered.

The classical formulation (Sec.~\ref{section:classical-ai}) evaluates the expected free energy of every policy in an enumerated policy space. This is exact with respect to the generative model, but the size of that space grows exponentially with the planning horizon, so in a state space of the present size only a single step of lookahead remains tractable. The dynamic programming scheme (Sec.~\ref{section:dpefe}) removes that limitation by evaluating expected free energy recursively backwards in time, which makes deep horizons computationally feasible. Its cost is a different one: a deep rollout is only as informative as the prior preference distribution $\mathbb{C}$ against which it is scored, so the scheme inherits whatever structure $\mathbb{C}$ does or does not possess. Counterfactual learning (Sec.~\ref{section:cl-method}) abandons forward evaluation altogether and instead learns a direct mapping from states to actions out of recent experience. It requires neither a rollout nor a preference distribution, but it can only reproduce regularities it has already encountered, and it has no mechanism for anticipating states it has not yet visited.

The three schemes therefore trade lookahead against tractability, and against dependence on the preference structure. Which of these trade-offs matters most is a property of the task rather than of the algorithms: an environment that rewards anticipation of unvisited states should favour the planning schemes, whereas one in which the currently observed state largely determines the correct action should favour the direct mapping. Sec.~\ref{sec:nte} then introduces the measure we use to compare the internal parameters of all three on a common footing, and Sec.~\ref{sec:stats} sets out the statistical framework used throughout.
}

\subsection{AIF: Classical formulation of decision-making in active inference}
\label{section:classical-ai}

\change{Active inference is a first-principle approach for modelling behaviour and revolves around a central postulate: every `agent' ensures survival by minimising observations ($o$) that are high in `entropy'}{Active inference is formulated around a central postulate, namely that an agent maintains its integrity by minimising the occurrence of observations ($o$) that are high in `entropy'} \cite{Friston2010}. This definition calls for a method to evaluate the entropy of an incoming observation ($o$). The entropy of observation, also known as `surprise' in the active inference literature \cite{FRISTON20231}, is defined as:
\begin{equation}\label{eq:surprise}
    S(o) = -\text{log}(P(o)).
\end{equation}

Recent experimental works \cite{Kagan2022, vincent_cell_2024, Isomura2023} use the concept of `surprise' in the active inference literature to design feedback signals for embodied agents that demonstrate improved learning with time. In Eq. \ref{eq:surprise}, $P(o)$ represents the probability of a given observation $o$. It should be noted that an agent is ignorant of the actual likelihood ($P_{true}(o)$) of any observation, as it is not in full control of the environment. The agent is not directly informed of how probable an incoming observation is. Hence, to estimate $P(o)$, the agent must internally encode and use a model to learn and predict the probability of incoming observations. This internal model is called the generative model \cite{Friston2012a, DaCosta2020}.

Traditionally, planning and decision-making by active inference agents revolve around minimising the variational free energy of the observations one expects in the \textit{future}. To implement this, we define a policy space comprising sequences of actions in time. The policy space in classical active inference \citep{Sajid2021} is defined as a collection of policies $\pi_n$:
\begin{equation}
    \Pi = \{\pi_{1}, \pi_{2}, ..., \pi_{N}\},
\end{equation}
which are themselves sequences of actions indexed in time; that is, $\pi = (u_{1}, u_{2}, ..., u_{T})$, where $u_t$ is one of the available actions in $U$, and $T$ is the agent's planning horizon. $N$ is the total number of unique policies defined by permutations of available actions $u$ over a time horizon of planning $T$.

To enable goal-directed behaviour, we must quantify the agent's preference for sample observations $o$. The prior preference for observations is usually defined as a categorical distribution over observations,
\begin{equation}
    \mathbb{C} = \mathrm{Cat}(o).
\end{equation}

So, if the value corresponding to an observation in $\mathbb{C}$ is the highest, it is the most preferred observation for the agent. 
Given these two additional parameters ($\Pi$ and $\mathbb{C}$), we can define a new quantity called the expected free energy (EFE) of a policy $\pi$ similar to the definition in \citep{Sajid2021, Parr2019a} as,

\begin{equation}\label{EFE_CAI_eqn}
    G(\pi) = \sum_{t=1}^T \underbrace{D_{KL}\left[ Q(o_{t} \vert \pi^t)  ~\vert \vert~ \mathbb{C} \right]}_\text{Risk} + \underbrace{\mathbb{E}_{Q(s_{t} \vert s_{t-1}, \pi^{t-1})} \left[ \mathbb{H}\left[ P(o_{t} \vert s_{t})\right] \right]}_\text{Expected ambiguity}.
\end{equation}

In Eq.\eqref{EFE_CAI_eqn} above, $\pi^t$ is the t-th element in $\pi$, i.e. the action corresponding to time $t$ for policy $\pi$. The term, $Q(o_{t} \vert \pi^t)$ represents the most likely observation caused by the policy $\pi$ at time $t$. $D_{KL}$ stands for the KL-divergence, which, when minimised, forces the distribution $Q(o_{t} \vert \pi^t)$ closer towards $\mathbb{C}$. This term is also called the "Risk," which measures the risk of an action that leads to outcomes that diverge from the agent's preferred states. Thus, actions are selected based on their expected KL divergence — essentially choosing actions that minimize this divergence, thereby minimizing "risk. This represents the goal-directed behaviour of the agent. The KL-divergence between two distributions, $P$ and $Q$, is defined as:
\begin{equation}
    D_{KL}(P \vert \vert Q) = \sum_{i} P(i) \text{log} \frac{P(i)}{Q(i)},
\end{equation}
and $P=Q$ if and only if $D_{KL}(P \vert \vert Q) = 0$.

In the second term of Eq.\ref{EFE_CAI_eqn}, $\mathbb{H}\left[ P(o_{t} \vert s_{t}) \right]$ stands for the (Shannon) entropy of $P(o_{t} \vert s_{t})$ defined as, 
\begin{equation}
    \mathbb{H}(P(o)) = -\sum_{o \epsilon O} P(o)\text{log}P(o).
\end{equation}

The second term is also called the `Expected ambiguity' term. When the expected entropy of $P(o_{t} \vert s_{t})$ w.r.t the belief $Q(s_{t} \vert s_{t-1}, \pi^{t-1})$ is lower, the agent is more confident of the state-observation mapping (i.e., $\mathbb{A}$) in its generative model. 

It may already be evident that the above formulation has one fundamental limitation: in the stochastic control problems commonly encountered in practice, the size of possible action spaces, $U$, and the time horizons of planning, $T$, make the policy space too large to be computationally tractable. For example, with eight available actions in $U$ and a time horizon of planning $T=15$, the total number of (definable) policies that need to be considered are ($3.5*10^{13}$) $35$ trillion. Even for this relatively small-scale example, this policy space is not computationally tractable to simulate agent behaviour (unless additional decision tree search methods \citep{fountasDeepActiveInference2020,championBranchingTimeActive2021a,championBranchingTimeActive2021} or policy amortisation \cite{fountasDeepActiveInference2020,Catal2020} are considered or by eliminating implausible policy trajectories using Occam's principle). In this work, we only consider one-step planning as the problem dimensions are intractable for planning with this approach. We refer to the agent using this method for planning as `AIF-1' abbreviating `Active inference agent with one-step planning'.

The limitation just described is one of tractability rather than of principle: the quantity being evaluated is the right one, but it cannot be evaluated over a policy space of this size. We now describe a scheme that uses dynamic programming principles \cite{dp_bellman_1966} to redefine the policy space and planning altogether, so that the same quantity can be computed at depth. In doing so it shifts the burden from the size of the policy space onto the prior preference distribution $\mathbb{C}$, which becomes the sole carrier of goal-directedness once policies are no longer enumerated.

\subsection{DPEFE: Dynamic programming in expected free energy}
\label{section:dpefe}
In a recent work \cite{Paul2021}, an efficient planning algorithm was proposed using dynamic programming principles. The algorithmic formulation generalised for a POMDP setting can be found in \cite{Paul2023}. We summarise this method below:

For a planning horizon of $T$ (i.e., the agent aims to reach goal state at time $T$), the EFE of the (last) action for the $T-1$th time step can be written as:

\begin{equation} \label{efeeqnpomdp}
    G(u_{T-1}, o_{T-1}) = D_{KL} [Q(o_{T} ~\vert~ u_{T-1},s_{T-1}) ~\vert \vert~ \mathbb{C}].
\end{equation}

The term, $G(u_{T-1} \vert o_{T-1})$ is the expected free energy associated with any action $u_{T-1}$, given we are in (hidden) state  $s_{T-1}$. This estimate measures how much we believe that the observations at time $T$ will align with our prior preference $\mathbb{C}$.

To estimate $Q(o_{T} \vert u_{T-1},s_{T-1})$, we make use of the prediction about states that can occur at time $T$, $Q(s_{T})$:

\begin{equation}\label{eq_g}
    Q(s_{T}~ \vert~ u_{T-1} ,s_{T-1})=\mathbb{B}_{u_{T-1}} \cdot Q(s_{T-1}),
\end{equation}

and given the prediction $Q(s_{T})$, we get 
\begin{align}\label{Eq.obs}
    Q(o_{T} ~\vert~ u_{T-1},s_{T-1})=\mathbb{A} \cdot Q(s_{T}~ \vert~ u_{T-1} ,s_{T-1}) \\
    = \mathbb{A} \cdot \left( \mathbb{B}_{u}^{T-1} \cdot Q(s_{T-1}) \right).
\end{align}

Next, using Eq.\ref{eq_g}, the corresponding action distribution (for action selection) is calculated at time $T$,
\begin{equation}\label{actdist}
    Q(u_{T-1} \vert o_{T-1})=\sigma \left( - G(u_{T-1} \vert o_{T-1}) \right).
\end{equation}

Here, we recursively calculate the expected free energy for actions and the corresponding action-distributions for time steps $T-2, T-3, ..., t=1$ backwards in time,\footnote{For times other than $T-1$, the first term in Eq.\ref{eq.rec.pl} does not contribute to solving the particular instance if $\mathbb{C}$ only accommodates preference to a (time-independent) goal-state. However, for a temporally informed $\mathbb{C}$, i.e. with a separate preference for reward maximisation at each time step, this term will meaningfully influence action selection.}

\begin{equation}\label{eq.rec.pl}
    G(u_{t} \vert o_{t})= \underbrace{D_{KL} [Q(o_{t+1} ~\vert~ u_{t},s_{t}) ~\vert \vert~ \mathbb{C}]}_\text{EFE of action at time $t$} + \underbrace{\mathbb{E}_{Q(o_{t+1},u_{t+1} \vert o_{t},u_{t})}[ G(u_{t+1} \vert o_{t+1})]}_\text{EFE of next action at $t+1$}.
\end{equation}

In the equation above, the second term condenses information about all future observations rather than doing a forward tree search in time. 
To inform $G(u_{t} \vert o_{t})$, we consider all possible observation-action pairs that can occur in time $t+1$ and use the previously evaluated $G(u_{t+1} \vert o_{t+1})$.
In Eq.\ref{eq.rec.pl}, we evaluate $Q(o_{t+1},u_{t+1} \vert o_{t},u_{t})$ using,
\begin{equation}\label{eq_q-split}
    Q(s_{t+1},u_{t+1} \vert s_{t},u_{t})= \underbrace{Q(s_{t+1} \vert s_{t},u_{t})}_\text{$\mathbb{B}$} \cdot \underbrace{Q(u_{t+1} \vert s_{t+1})}_\text{Action distribution}.
\end{equation}
We then map the distribution $Q(s_{t+1},u_{t+1} \vert s_{t},u_{t})$ to the observation space and evaluate $Q(o_{t+1},u_{t+1} \vert o_{t},u_{t})$ using the likelihood mapping $\mathbb{A}$.
In Eq.\ref{eq_q-split}, we assume that actions in time are independent of each other, i.e. $u_{t}$ is independent of $u_{t+1}$. Even though actions are assumed to be explicitly independent in time, the information (and hence desirability) about actions is also informed backwards from the recursive evaluation of expected free energy. 

While evaluating the EFE, $G$, backwards in time, we used the action distribution in Eq.\ref{actdist}. This action distribution can be directly used for action selection.
Given an observation $o$ at time $t$, $u_{t}$ may be sampled \footnote{Precision of action-section may be controlled by introducing a positive constant inside the softmax function $\sigma(.)$ in Eq.\ref{actdist}. The higher the constant, the higher the chance of selecting the action with less EFE.} from, 
\begin{equation}\label{eqn:action_sampling}
    u_{t} \sim Q(u_{t} \vert o_{t} = o).
\end{equation}

We refer to agents using the DPEFE method for decision making as `DP-T', abbreviating `Active inference agent using dynamic programming with planning horizon $T$.

\subsection{CFL: Counterfactual learning}
\label{section:cl-method}

Both schemes described so far evaluate the future before acting, and both depend on the prior preference distribution $\mathbb{C}$ to say which futures are desirable. A different approach dispenses with each of these commitments. Rather than planning, it relies on counterfactual learning (CFL), in which goal-directedness enters through a risk signal attached to recent experience rather than through a preference over observations. In the counterfactual learning method, the agent learns a state-action mapping $\mathbb{CL}$ instead of evaluating EFE directly \footnote{For the exact form of the generative model and free energy, refer to \cite{Isomura2020}}. This state-action mapping is learned using a 'Risk' parameter $\Gamma(t)$ using the update equation as given in \cite{Isomura2020} as:

\begin{equation}
    \mathbb{CL} \leftarrow \mathbb{CL} + t ~ \langle(1 - 2 ~\Gamma(t)) \langle u_{t} \otimes s_{t-1} \rangle\rangle.
    \label{cp-clmodeleqn}
\end{equation}

Here, $\langle \cdot \rangle$ refers to the average over time, and $\otimes$ is the Kronecker-product operator.
Given the state-action mapping $\mathbb{CL}$, agent samples actions from the distribution, 
\begin{equation}
    P(u \vert s_{t-1})_{CL} = \sigma \left( \ln~ \mathbb{CL} \cdot s_{t-1} \right).
    \label{cl-action-selection-eqn}
\end{equation}

\change{The free parameter in our model is the number of past instances (of state-action pairs) the agent stores in memory use in every time-step to learn $\mathbb{CL}$ in the update equation above. In this paper, `CFL-T' represents the active inference agent with a memory horizon of $T$.}{The free parameter in our model is the number of past instances used at each time step to update $\mathbb{CL}$. We refer to this quantity as the \textit{finite state-action update window} of length $T$, and write `CFL-$T$' for the corresponding agent. The term `memory horizon' is equalently used and should be understood in the strictly algorithmic sense set out below.}

Algorithmically, the window is a first-in-first-out buffer holding the most recent pairs $(Q(s_{\tau}), u_{\tau})$, i.e. the agent's posterior belief over the hidden state at time $\tau$ together with the action it sampled at that time. The buffer therefore stores state-action pairs, and not raw observations or hidden states in isolation. Once it is full, the update in Eq.\ref{cp-clmodeleqn} is applied at every subsequent time step, once for each buffered pair, with a separate value of the risk term $\Gamma$ maintained for each position in the buffer. Because the likelihood $\mathbb{A}$ is an identity mapping in this task (Sec.\ref{section:pomdp}), $Q(s_{\tau})$ is concentrated on the observed state, so in practice the window reduces to a buffer of observed state-action pairs.

Two qualifications follow, and we state them here because they bear on how the results should be read. First, this window is an algorithmic update rule, and not an independently measured biological memory mechanism. Nothing in our simulations establishes that neuronal cultures maintain a comparable buffer, and we make no such claim. Second, Pong under a fully observable state representation is not a strong assay of memory in the cognitive or biological sense, precisely because the observation at each step already encodes the information required to act. Larger values of $T$ should therefore be read as increasing the amount of recent experience folded into each update, rather than as evidence for long-range recall in vitro. We return to this point in Sec.\ref{sec:discussions}.

The functional form of $\Gamma(t)$ used in the simulations of this work is:
\begin{equation}
    \Gamma(t)_{prior} = 0.55 
    \label{gamma-prior-eqn}
\end{equation}

A value of $\Gamma$ greater than 0.55 corresponds to ``higher risk'' in the CFL method. An initial value greater than 0.5 is necessary to enable learning \cite{Isomura2020}.

For updating $\Gamma$, we use the equation,
\begin{equation}
    \Gamma(t) \leftarrow \Gamma(t) - \frac{1}{T_{goal} - t}.
    \label{gamma-update-eqn}
\end{equation}

\add{One implementation detail of this update affects behaviour and is not implied by the equations above: the counterfactual update is applied only once the buffer holds $T$ state-action pairs. For the first $T$ steps of an episode no update is made, so a shorter memory horizon begins updating sooner within each episode as well as averaging over fewer pairs.} 
Here, $T_{goal}$ is when the agent receives a positive environmental reward. So, the sooner the agent reaches the `goal-state', the quicker the $\Gamma(t)$, i.e., risk converges to zero. All the update rules this paper defines can be derived from the postulate that the agent tries to minimise the (variational) free energy w.r.t the generative model \cite{Isomura2023, Paul2023}. It was observed in \cite{Paul2024} and \cite{khajehnejad2024biologicalneuronscompetedeep} that this variation of the CFL method is well suited for environments that require spontaneous decision-making and is data-efficient (Environments like the Cart Pole (OpenAIGym) \cite{Barto1983}, game of Pong \cite{khajehnejad2024biologicalneuronscompetedeep}).

The algorithms implemented in this paper for decision-making (DPEFE \cite{Paul2023} and CFL \cite{Paul2023}) are low-cost computational methods against classical active inference algorithms. The improvement in computational complexity of these models enabled simulations in the state space consisting of more than two thousand states\change{ and has the potential to scale to more demanding environments.}{.}

\subsection{Predictive information of the learned transition model}
\label{sec:predictive-depth}

\new{
The planning schemes (DPEFE-T, AIF-T etc.) evaluate policies by rolling the transition parameters $\mathbb{B}$ forward and scoring the resulting predictions. How far that is worth doing depends on how long a rollout remains informative, which is a property of the agent's estimate rather than of the environment. To measure it, we form the $h$-step operator by composing the fitted transition matrix with itself. Then, we take the entropy of each of its columns, i.e. of the predicted distribution over states $h$ steps after starting from a known state. We then it normalise by $\log N$, the entropy of the uniform distribution over the same support. A value of one means the rollout carries no information about where the ball will be.

We compute this from the transition parameters of the DP-5 agent on three seeds, averaging over the three actions and over all starting states. Retaining a full transition matrix per seed is expensive, so this is a demonstration of the mechanism rather than an estimate over the population of seeds, and we treat it accordingly. Results are summarised in Supplementary Table \ref{tab:planningdepth}.
}

\subsection{Matched settings across the three schemes}
\label{sec:priorcheck}

\new{
The three schemes share the generative model set out in Sec.~\ref{section:exp_genmodel}, and they also share the settings that are not intrinsic to a scheme. Supplementary Table \ref{tab:hyperparams} lists these: the action precision governing the softmax over policies, the learning rate on the transition parameters, the numerical floor, and the number of episodes per trial are identical in all three. What differs between them is the horizon, which is the quantity under study.
}
\new{
To observe the effects of different prior preference, initially the classical formulation drew its prior preferences $\mathbb{C}$ and its initial state prior $\mathbb{D}$ from a random categorical distribution, whereas the other two schemes initialise both as uniform. Because a poorly matched preference distribution would disadvantage a planning scheme for reasons that have nothing to do with planning. We re-ran the classical agent with $\mathbb{C}$ and $\mathbb{D}$ set to uniform, holding everything else fixed, and compared the two specifications seed by seed on twelve seeds. The paired difference in second-window rally length is $-0.056$ ($95\%$ CI $[-0.177, 0.065]$; paired $t = -0.909$, $p = 0.38$; Wilcoxon $p = 0.57$), i.e. uniform priors are marginally worse rather than better, and are not distinguishable from the published specification. Each of the twelve seeds was run under both specifications, and for each seed we took the mean rally length (hits per rally) over the second window (6--20 min) and subtracted the value obtained with the published random $\mathbb{C}$ and $\mathbb{D}$ from the value obtained with uniform ones, so a negative difference means that uniform priors performed worse. On these seeds the mean second-window rally length was $0.70$ hits with uniform priors and $0.76$ hits with the published specification, both below the cultures ($0.85$ for MCC and $0.88$ for HCC), and the confidence interval of the difference includes zero. With twelve seeds the interval still admits differences of up to about $0.12$ hits in either direction, so the comparison shows that the choice of prior does not account for the low performance of AIF-1, rather than that the two specifications are equivalent. The limitation of AIF-1 in this task is therefore its one-step horizon and not its preference specification, and we have retained the published runs.

The full simulation protocol, i.e. the software environment, the seeding procedure, the trial and episode structure, and the rule by which the two time windows are defined, is set out in Supplementary Methods.}

\subsection{Observation noise and unreliable action execution}
\label{sec:noise-methods}

\new{
The likelihood of our generative model is the identity, so an observation determines the state and the task is fully observable. To establish whether our results depend on that idealisation we relax it in two independent ways.

Observation noise mixes the likelihood towards uniform, $\mathbb{A}_{\varepsilon} = (1-\varepsilon)\mathbb{A} + \varepsilon/n$, where $n$ is the number of observations in a modality, so that larger $\varepsilon$ makes an observation progressively less informative about the state. Unreliable action execution resamples the executed action uniformly with probability $\eta$, applied after the agent has selected its action, so the agent continues to believe it acted as it chose while the environment executed something else; this models an imperfect readout of the action rather than a change of policy. Both manipulations are no-ops at zero, so $\varepsilon = \eta = 0$ is a control condition that must recover the fully observable conditions and we verified it numerically. The sample sizes differ by agent. CFL-4 is run at the published size of one hundred seeds per condition. DP-5 costs some three quarters of an hour per seed and is run at six, which is a pilot rather than a powered comparison.}

\subsection{Baselines outside active inference}
\label{sec:baseline-methods}

\new{
To establish whether the behaviour we report depends on active inference specifically, we compare the agents above with four controllers that fall outside the framework. All four interact with the environment through the same loop as the active inference agents, using the same reset protocol, the same definition of a hit, the same episode termination rule and the same window-splitting rule, so that the resulting data enter the analysis on identical terms. Each is run for $99$ seeds of $70$ episodes, matching the sample size of the DPEFE and AIF-1 agents.

\begin{itemize}
    \item \textbf{Random.} Actions are drawn uniformly from the three available at every time step. The controller has no state and cannot learn, so it fixes the floor against which any claim of learning must be read, and it calibrates the improvement metric defined in Sec.~\ref{sec:stats}.
    \item \textbf{Heuristic.} A reactive controller that compares the vertical position of the ball with that of the paddle and moves the paddle towards the ball, remaining still when the two coincide. It has full state access but no memory and no learning. It establishes what the task itself demands.
    \item \textbf{Dyna-Q.} Tabular Dyna-Q over the same $2432$ states and three actions, with $\epsilon = 0.1$, $\alpha = 0.5$ and $\gamma = 0.9$. The reward mirrors the risk signal available to the counterfactual learning agent, i.e. $+1$ on a hit, $-1$ when an episode ends in a miss, and $0$ otherwise. This is generic model-based reinforcement learning with none of the active inference machinery, and it tests whether the advantage we report is specific to the framework or follows from short-term memory together with feedback.
    \item \textbf{Dyna-Q with replay.} The same agent with ten replay updates per transition, giving a stronger tabular baseline at equal experience.
\end{itemize}

The baseline contrasts are corrected within their own family of comparisons, separately from the agent--culture contrasts reported in Sec.~\ref{sec:cl_result} and Sec.~\ref{sec:add_results}. They address a different question, and pooling the two families would alter the correction applied to the contrasts on which the main analysis rests.
}

\subsection{Explainability: Normalised total entropy}
\label{sec:nte}

The three schemes expose different parameters to inspection: $\mathbb{C}$ and $\mathbb{B}$ in the two planning schemes, and the $\mathbb{CL}$ mapping together with the risk term $\Gamma$ in counterfactual learning. To compare them on a common footing we need a single summary of how informed a probability-encoding parameter has become. To that end, we define the total entropy (TE) of a parameter as,
\begin{equation}
\label{eq:nme}
    \text{TE}(Z) = \sum_{X \epsilon Z} \mathbb{H}(X) = -\sum_{X \epsilon Z} \sum_{x \epsilon X} p(x) \text{log}~p(x) .
\end{equation}

In Eq.\ref{eq:nme}, $Z$ is a parameter like the $\mathbb{CL}$ mapping in the counterfactual learning method, and $\mathbb{H}$, the Shannon entropy. $X$ is a probability distribution collection comprising the parameter $Z$. For example, the $\mathbb{CL}$ mapping is made up of probability distributions $P(u | s_{t}) ~ \forall ~s \epsilon S$, representing how probable an action $u$ is, given a state $s$ at time $t$.

\new{
Total entropy is bounded above, and the bound is available in closed form: a parameter is least informed when every distribution it contains is uniform. For a parameter $Z$ comprising distributions over supports of size $|X|$, this maximum is
\begin{equation}
\label{eq:nte-max}
    \text{TE}_{\max}(Z) = \sum_{X \epsilon Z} \text{log}~|X| .
\end{equation}
We therefore define the normalised total entropy (NTE) as the ratio
\begin{equation}
\label{eq:nte}
    \text{NTE}(Z) = \frac{\text{TE}(Z)}{\text{TE}_{\max}(Z)} ,
\end{equation}
which takes the value one when the parameter carries no information about the task, and approaches zero as it becomes fully determined. For the generative model used here, $\text{TE}_{\max}$ is $2671.83$ nats for the $\mathbb{CL}$ mapping, which holds one distribution over three actions for each of $2432$ states; $414.68$ nats for the ball-x factor of $\mathbb{B}$, which holds one distribution over $38$ successor states for each state-action pair; and $16.64$ nats for each of the ball-y and paddle-y factors, which are not controllable.

Normalising against the analytic maximum, rather than against the largest value observed within a run, is what makes NTE comparable across agents, parameters and time. A within-run scaling sets every agent to one at the point where it happens to be least informed, and therefore reports the shape of a trace rather than its level. Raw total entropies, together with the maxima above, are given in Supplementary Table \ref{tab:nte}.
}

\subsection{Statistical analysis}
\label{sec:stats}

\new{
\paragraph{Units of analysis}
We distinguish three levels throughout. An \textit{episode} is a single Pong rally. A \textit{session} is one twenty-minute trial, corresponding in-vitro to a (chip, date, session) triple and in-silico to one random seed. A \textit{replicate} is the level at which observations may be treated as independent: in-vitro this is the culture, i.e. the chip, because several sessions are typically recorded from the same culture, and in-silico it is the seed, because separate runs share no state. Table \ref{tab:units} reports these counts. The episode counts are matched by design, but the biological sessions are nested within a small number of cultures, and treating sessions as independent therefore understates biological variability. Resampling whole cultures widens the $95\%$ interval on relative improvement by a factor of $1.97$ for HCC and $1.73$ for MCC, while leaving the in-silico intervals essentially unchanged.
}

\begin{table}[ht]
\centering
\caption{Units of analysis. Episode counts are matched by design between in-vitro and in-silico groups, but the biological sessions are nested within a small number of cultures.}
\label{tab:units}
\begin{tabular}{llrrr}
\toprule
Source & Group & Replicates & Sessions & Episodes \\
\midrule
In vitro & MCC & 7 chips & 110 & 7,317 \\
In vitro & HCC & 12 chips & 138 & 9,528 \\
In silico & CFL-$T$ (each $T$) & 100 seeds & 100 & 7,000 \\
In silico & AIF-1 & 99 seeds & 99 & 6,930 \\
In silico & DP-$T$ (each $T$) & 99 seeds & 99 & 6,930 \\
\bottomrule
\end{tabular}
\\[4pt]\footnotesize{Replicates are the level at which observations are independent: the culture (chip) in vitro, the random seed in silico. Because many sessions are recorded from the same culture, treating sessions as independent understates biological variability; resampling whole chips widens the 95\% interval on relative improvement by 1.99$\times$ for HCC and 1.73$\times$ for MCC, while leaving in-silico intervals unchanged.}
\end{table}

\new{
\paragraph{Behavioural metrics}
From each session we derive three metrics, which do not share a common directionality. The average rally length is the mean hit count over episodes, and higher values are better. The percentage of long rallies is the proportion of episodes with at least three hits, and higher values are better. The percentage of aces is the proportion of episodes with no hits, and \textit{lower} values are better. Relative improvement is not a composite of these three. It is computed from the rally length alone, as
\begin{equation}\label{eq:ri}
    \mathrm{RI} = 100 \times \frac{\overline{r}_{\,6\text{-}20} - \overline{r}_{\,0\text{-}5}}{\overline{r}_{\,0\text{-}5}},
\end{equation}
where $\overline{r}_{\,0\text{-}5}$ and $\overline{r}_{\,6\text{-}20}$ are the mean rally lengths in the first and second time windows of a session. Sessions whose first window has a mean of exactly zero are excluded, as RI is undefined for them; this removes one CFL-4 seed and no biological session. It should be noted that Eq.\ref{eq:ri} averages per-session ratios, and the mean of ratios is not the ratio of means. The measure is therefore sensitive to sessions with a small first-window denominator; for AIF-1 this inverts the sign, giving a positive mean relative improvement ($+15.8\%$) alongside a negative absolute change ($-0.016$).

For this reason we take the \textit{absolute} performance reached in the second window as the primary basis for comparison between agents and cultures, and report relative improvement as a secondary measure, for continuity with the original DishBrain analysis. Two further observations support this choice. Relative improvement cannot separate the two culture types from each other (HCC $-$ MCC $= +14.1$ percentage points, $95\%$ CI $[-20.9, 45.9]$), which sets a resolution floor for any comparison made with it; and no agent--culture contrast on that measure reaches statistical equivalence (Supplementary Table \ref{tab:contrastsri}). Absolute rally length, by contrast, resolves 13 of the 18 contrasts (Table \ref{tab:contrasts}).
}

\section{Acknowledgments}
A.R. received funding from the Australian Research Council (refs.: DP200100757, FT250100563 and DP260104251), the Australian National Health and Medical Research Council (investigator grant: 1194910). A.R. and B.J.K. received funding from the Australian Office of National Intelligence (ONI) through their National Intelligence and Discovery Grant scheme (ref.: NI230100153). A.R. is a CIFAR Global Scholar in the Brain, Mind \& Consciousness Program and acknowledges the support funding as part of CIFAR Next-Generation Initiatives (ref.: GS22-009) and their Catalyst grants (refs.: CF-0224 and CF-0491). B.J.K. \& F.H. are employees of and hold options or shares in Cortical Labs Pty Ltd.

\section{Code Availability and AI Use Declaration}
The associated code is available in the public repository: \textit{https://github.com/aswinpaul/adaptive-behaviour}. The first author declares the use of Claude Opus 5 as programming and writing assistants.

\section{Author credit statement}

\begin{itemize}
    \item Aswin Paul: Conceptualization, Methodology, Software, Validation, Formal analysis, Investigation, Writing - Original Draft, Writing - Review and Editing
    \item Moein Khajehnjad: Writing - Review and Editing
    \item Forough Habibollahi: Writing - Review and Editing
    \item Brett J Kagan: Writing - Review and Editing
    \item Adeel Razi: Conceptualization, Methodology, Supervision, Project administration, Funding acquisition, Writing - Review and Editing
\end{itemize}

\bibliographystyle{unsrtnat}
\bibliography{bib_aif}

@article{
khajehnjad2025,
author = {Moein Khajehnejad  and Forough Habibollahi  and Alon Loeffler  and Aswin Paul  and Adeel Razi  and Brett J. Kagan },
title = {Dynamic Network Plasticity and Sample Efficiency in Biological Neural Cultures: A Comparative Study with Deep Reinforcement Learning},
journal = {Cyborg and Bionic Systems},
volume = {6},
number = {},
pages = {0336},
year = {2025},
doi = {10.34133/cbsystems.0336},
URL = {https://spj.science.org/doi/abs/10.34133/cbsystems.0336},
eprint = {https://spj.science.org/doi/pdf/10.34133/cbsystems.0336}}

@article{habibollahi2023critical,
  title={Critical dynamics arise during structured information presentation within embodied in vitro neuronal networks},
  author={Habibollahi, Forough and Kagan, Brett J and Burkitt, Anthony N and French, Chris},
  journal={Nature Communications},
  volume={14},
  number={1},
  pages={5287},
  year={2023},
  publisher={Nature Publishing Group UK London}
}

@article{kagan2023technology,
  title={The technology, opportunities and challenges of synthetic biological intelligence},
  author={Kagan, Brett J and Gyngell, Christopher and Lysaght, Tamra and Cole, Victor M and Sawai, Tsutomu and Savulescu, Julian},
  journal={Biotechnology advances},
  pages={108233},
  year={2023},
  publisher={Elsevier}
}

@article{neftci2019reinforcement,
  title={Reinforcement learning in artificial and biological systems},
  author={Neftci, Emre O and Averbeck, Bruno B},
  journal={Nature Machine Intelligence},
  volume={1},
  number={3},
  pages={133--143},
  year={2019},
  publisher={Nature Publishing Group}
}

@ARTICLE{Evans2024-dna,
  title    = "Pattern recognition in the nucleation kinetics of non-equilibrium
              self-assembly",
  author   = "Evans, Constantine Glen and O'Brien, Jackson and Winfree, Erik
              and Murugan, Arvind",
  journal  = "Nature",
  volume   =  625,
  number   =  7995,
  pages    = "500--507",
  month    =  jan,
  year     =  2024
}

@Article{vincent_cell_2024,
author={Strong, Vincent
and Holderbaum, William
and Hayashi, Yoshikatsu},
title={Electro-active polymer hydrogels exhibit emergent memory when embodied in a simulated game environment},
journal={Cell Reports Physical Science},
year={2024},
month={2024/08/28},
publisher={Elsevier},
issn={2666-3864},
doi={10.1016/j.xcrp.2024.102151},
url={https://doi.org/10.1016/j.xcrp.2024.102151}
}

@Misc{friston2023supervised,
  title={Supervised structure learning},
  author={Friston, Karl J and Da Costa, Lancelot and Tschantz, Alexander and Kiefer, Alex and Salvatori, Tommaso and Neacsu, Victorita and Koudahl, Magnus and Heins, Conor and Sajid, Noor and Markovic, Dimitrije and others},
  journal={arXiv preprint arXiv:2311.10300},
  year={2023}
}

@misc{albarracin2023-xaif,
      title={Designing explainable artificial intelligence with active inference: A framework for transparent introspection and decision-making}, 
      author={Mahault Albarracin and Inês Hipólito and Safae Essafi Tremblay and Jason G. Fox and Gabriel René and Karl Friston and Maxwell J. D. Ramstead},
      year={2023},
      eprint={2306.04025},
      archivePrefix={arXiv},
      primaryClass={cs.AI},
      url={https://arxiv.org/abs/2306.04025}, 
}

@article{friston2018deep,
  title={Deep temporal models and active inference},
  author={Friston, Karl J and Rosch, Richard and Parr, Thomas and Price, Cathy and Bowman, Howard},
  journal={Neuroscience \& Biobehavioral Reviews},
  volume={90},
  pages={486--501},
  year={2018},
  publisher={Elsevier}
}

@ARTICLE{Barto1983,
  author={Barto, Andrew G. and Sutton, Richard S. and Anderson, Charles W.},
  journal={IEEE Transactions on Systems, Man, and Cybernetics}, 
  title={Neuronlike adaptive elements that can solve difficult learning control problems}, 
  year={1983},
  volume={SMC-13},
  number={5},
  pages={834-846},
  doi={10.1109/TSMC.1983.6313077}}

@misc{khajehnejad2024biologicalneuronscompetedeep,
      title={Biological Neurons Compete with Deep Reinforcement Learning in Sample Efficiency in a Simulated Gameworld}, 
      author={Moein Khajehnejad and Forough Habibollahi and Aswin Paul and Adeel Razi and Brett J. Kagan},
      year={2024},
      eprint={2405.16946},
      archivePrefix={arXiv},
      primaryClass={q-bio.NC},
      url={https://arxiv.org/abs/2405.16946}, 
}

@Article{Friston2023inductiveinf,
  author  = {Karl J. Friston and Tommaso Salvatori and Takuya Isomura and Alexander Tschantz and Alex Kiefer and Tim Verbelen and Magnus T. Koudahl and Aswin Paul and Thomas Parr and Adeel Razi and Brett J. Kagan and Christopher L. Buckley and Maxwell James D. Ramstead},
  journal = {ArXiv},
  title   = {Active Inference and Intentional Behaviour},
  year    = {2023},
  volume  = {abs/2312.07547},
  url     = {https://api.semanticscholar.org/CorpusID:266191299},
}

@Article{Isomura2020,
  author   = {Isomura, Takuya and Friston, Karl},
  journal  = {Neural Computation},
  title    = {{Reverse-Engineering Neural Networks to Characterize Their Cost Functions}},
  year     = {2020},
  issn     = {0899-7667},
  month    = {11},
  number   = {11},
  pages    = {2085-2121},
  volume   = {32},
  doi      = {10.1162/neco_a_01315},
  eprint   = {https://direct.mit.edu/neco/article-pdf/32/11/2085/1865423/neco\_a\_01315.pdf},
  url      = {https://doi.org/10.1162/neco\_a\_01315},
}

@Article{Friston2010,
  author   = {Friston, Karl},
  title    = {The free-energy principle: a unified brain theory?},
  doi      = {10.1038/nrn2787},
  issn     = {1471-0048},
  number   = {2},
  pages    = {127--138},
  url      = {https://doi.org/10.1038/nrn2787},
  volume   = {11},
  journal  = {Nature Reviews Neuroscience},
  refid    = {Friston2010},
  year     = {2010},
}

@article{
dp_bellman_1966,
author = {Richard Bellman },
title = {Dynamic Programming},
journal = {Science},
volume = {153},
number = {3731},
pages = {34-37},
year = {1966},
doi = {10.1126/science.153.3731.34},
URL = {https://www.science.org/doi/abs/10.1126/science.153.3731.34},
eprint = {https://www.science.org/doi/pdf/10.1126/science.153.3731.34}}

@article{chat_gpt_2023,
title = {ChatGPT: Jack of all trades, master of none},
journal = {Information Fusion},
volume = {99},
pages = {101861},
year = {2023},
issn = {1566-2535},
doi = {https://doi.org/10.1016/j.inffus.2023.101861},
url = {https://www.sciencedirect.com/science/article/pii/S156625352300177X},
author = {Jan Kocoń and Igor Cichecki and Oliwier Kaszyca and Mateusz Kochanek and Dominika Szydło and Joanna Baran and Julita Bielaniewicz and Marcin Gruza and Arkadiusz Janz and Kamil Kanclerz and Anna Kocoń and Bartłomiej Koptyra and Wiktoria Mieleszczenko-Kowszewicz and Piotr Miłkowski and Marcin Oleksy and Maciej Piasecki and Łukasz Radliński and Konrad Wojtasik and Stanisław Woźniak and Przemysław Kazienko}
}

@article{gm_weather_2023,
title = {Development of robust detector using the weather deep generative model for outdoor monitoring system},
journal = {Expert Systems with Applications},
volume = {234},
pages = {120984},
year = {2023},
issn = {0957-4174},
doi = {https://doi.org/10.1016/j.eswa.2023.120984},
url = {https://www.sciencedirect.com/science/article/pii/S0957417423014860},
author = {Kyo-Hoon Jin and Kyung-Su Kang and Baek-Kyun Shin and June-Hyoung Kwon and Soo-Jin Jang and Young-Bin Kim and Han-Guk Ryu}
}

@Article{Ingraham2023,
  author   = {Ingraham, John B. and Baranov, Max and Costello, Zak and Barber, Karl W. and Wang, Wujie and Ismail, Ahmed and Frappier, Vincent and Lord, Dana M. and Ng-Thow-Hing, Christopher and Van Vlack, Erik R. and Tie, Shan and Xue, Vincent and Cowles, Sarah C. and Leung, Alan and Rodrigues, João V. and Morales-Perez, Claudio L. and Ayoub, Alex M. and Green, Robin and Puentes, Katherine and Oplinger, Frank and Panwar, Nishant V. and Obermeyer, Fritz and Root, Adam R. and Beam, Andrew L. and Poelwijk, Frank J. and Grigoryan, Gevorg},
  journal  = {Nature},
  title    = {Illuminating protein space with a programmable generative model},
  year     = {2023},
  issn     = {1476-4687},
  number   = {7989},
  pages    = {1070--1078},
  volume   = {623},
  doi      = {10.1038/s41586-023-06728-8},
  refid    = {Ingraham2023},
  url      = {https://doi.org/10.1038/s41586-023-06728-8},
}

@ARTICLE{gmm_ieee2005,
  author={Pernkopf, F. and Bouchaffra, D.},
  journal={IEEE Transactions on Pattern Analysis and Machine Intelligence}, 
  title={Genetic-based EM algorithm for learning Gaussian mixture models}, 
  year={2005},
  volume={27},
  number={8},
  pages={1344-1348},
  doi={10.1109/TPAMI.2005.162}}

@article{FRISTON20231,
title = {The free energy principle made simpler but not too simple},
journal = {Physics Reports},
volume = {1024},
pages = {1-29},
year = {2023},
note = {The free energy principle made simpler but not too simple},
issn = {0370-1573},
doi = {https://doi.org/10.1016/j.physrep.2023.07.001},
url = {https://www.sciencedirect.com/science/article/pii/S037015732300203X},
author = {Karl Friston and Lancelot {Da Costa} and Noor Sajid and Conor Heins and Kai Ueltzhöffer and Grigorios A. Pavliotis and Thomas Parr}
}

@Article{DaCosta2020,
  author   = {Da Costa, Lancelot and Parr, Thomas and Sajid, Noor and Veselic, Sebastijan and Neacsu, Victorita and Friston, Karl},
  title    = {Active inference on discrete state-spaces: A synthesis},
  doi      = {10.1016/j.jmp.2020.102447},
  issn     = {0022-2496},
  pages    = {102447},
  url      = {https://www.sciencedirect.com/science/article/pii/S0022249620300857},
  volume   = {99},
  journal  = {Journal of Mathematical Psychology},
  year     = {2020},
}

@Article{Parr2019a,
  author   = {Parr, Thomas and Friston, Karl J.},
  journal  = {Biological Cybernetics},
  title    = {Generalised free energy and active inference},
  year     = {2019},
  issn     = {1432-0770},
  number   = {5},
  pages    = {495--513},
  volume   = {113},
  doi      = {10.1007/s00422-019-00805-w},
  refid    = {Parr2019},
  url      = {https://doi.org/10.1007/s00422-019-00805-w},
}

@Article{Friston2012,
  author  = {Friston, Karl},
  journal = {Entropy (Basel, Switzerland)},
  title   = {A Free Energy Principle for Biological Systems},
  year    = {2012},
  month   = {11},
  pages   = {2100-2121},
  volume  = {14},
  doi     = {10.3390/e14112100},
}

@Article{Kaelbling1998,
  author   = {Leslie Pack Kaelbling and Michael L. Littman and Anthony R. Cassandra},
  journal  = {Artificial Intelligence},
  title    = {Planning and acting in partially observable stochastic domains},
  year     = {1998},
  issn     = {0004-3702},
  number   = {1},
  pages    = {99-134},
  volume   = {101},
  doi      = {https://doi.org/10.1016/S0004-3702(98)00023-X},
  url      = {https://www.sciencedirect.com/science/article/pii/S000437029800023X},
}

@Article{Friston2009,
  author    = {Friston, Karl J. AND Daunizeau, Jean AND Kiebel, Stefan J.},
  title     = {Reinforcement Learning or Active Inference?},
  doi       = {10.1371/journal.pone.0006421},
  number    = {7},
  pages     = {1-13},
  url       = {https://doi.org/10.1371/journal.pone.0006421},
  volume    = {4},
  journal   = {PLOS ONE},
  month     = {07},
  publisher = {Public Library of Science},
  year      = {2009},
}

@Article{Friston2012a,
  author   = {Friston, Karl and Samothrakis, Spyridon and Montague, Read},
  journal  = {Biological Cybernetics},
  title    = {Active inference and agency: optimal control without cost functions},
  year     = {2012},
  issn     = {1432-0770},
  number   = {8},
  pages    = {523--541},
  volume   = {106},
  doi      = {10.1007/s00422-012-0512-8},
  refid    = {Friston2012},
  url      = {https://doi.org/10.1007/s00422-012-0512-8},
}

@Article{Sajid2021,
  author    = {Sajid, Noor and Ball, Philip J. and Parr, Thomas and Friston, Karl J.},
  title     = {Active Inference: Demystified and Compared},
  doi       = {10.1162/neco_a_01357},
  issn      = {0899-7667},
  number    = {3},
  pages     = {674--712},
  url       = {https://doi.org/10.1162/neco_a_01357},
  volume    = {33},
  comment   = {doi: 10.1162/neco_a_01357},
  journal   = {Neural Computation},
  month     = jan,
  publisher = {MIT Press},
  year      = {2021},
}

@InProceedings{Catal2020,
  author    = {O. {Çatal} and T. {Verbelen} and J. {Nauta} and C. D. {Boom} and B. {Dhoedt}},
  booktitle = {ICASSP 2020 - 2020 IEEE International Conference on Acoustics, Speech and Signal Processing (ICASSP)},
  title     = {Learning Perception and Planning With Deep Active Inference},
  year      = {2020},
  pages     = {3952-3956},
  doi       = {10.1109/ICASSP40776.2020.9054364},
}

@InProceedings{Millidge2020a,
  author    = {Millidge, Beren and Tschantz, Alexander and Seth, Anil K. and Buckley, Christopher L.},
  booktitle = {Active Inference},
  title     = {On the Relationship Between Active Inference and Control as Inference},
  year      = {2020},
  address   = {Cham},
  editor    = {Verbelen, Tim and Lanillos, Pablo and Buckley, Christopher L. and De Boom, Cedric},
  pages     = {3--11},
  publisher = {Springer International Publishing},
  doi       = {https://link.springer.com/chapter/10.1007/978-3-030-64919-7_1},
  isbn      = {978-3-030-64919-7},
}

@Article{Lovejoy1991,
  author   = {Lovejoy, William S.},
  journal  = {Annals of Operations Research},
  title    = {A survey of algorithmic methods for partially observed Markov decision processes},
  year     = {1991},
  issn     = {1572-9338},
  number   = {1},
  pages    = {47--65},
  volume   = {28},
  doi      = {10.1007/BF02055574},
  refid    = {Lovejoy1991},
  url      = {https://doi.org/10.1007/BF02055574},
}

@InProceedings{Paul2021,
  author    = {Paul, Aswin and Sajid, Noor and Gopalkrishnan, Manoj and Razi, Adeel},
  booktitle = {Machine Learning and Principles and Practice of Knowledge Discovery in Databases},
  title     = {Active Inference for Stochastic Control},
  year      = {2021},
  address   = {Cham},
  pages     = {669--680},
  publisher = {Springer International Publishing},
  doi       = {https://doi.org/10.1007/978-3-030-93736-2_47},
  isbn      = {978-3-030-93736-2},
}

@article{fountasDeepActiveInference2020,
  title = {Deep Active Inference Agents Using {{Monte-Carlo}} Methods},
  author = {Fountas, Zafeirios and Sajid, Noor and Mediano, Pedro A. M. and Friston, Karl},
  year = {2020},
  month = jun,
  journal = {arXiv:2006.04176 [cs, q-bio, stat]},
  eprint = {2006.04176},
  primaryclass = {cs, q-bio, stat},
  urldate = {2020-07-16},
  archiveprefix = {arxiv}
}

@article{championBranchingTimeActive2021a,
  title = {Branching {{Time Active Inference}}: The Theory and Its Generality},
  shorttitle = {Branching {{Time Active Inference}}},
  author = {Champion, Th{\'e}ophile and Da Costa, Lancelot and Bowman, Howard and Grze{\'s}, Marek},
  year = {2021},
  month = nov,
  journal = {arXiv:2111.11107 [cs]},
  eprint = {2111.11107},
  primaryclass = {cs},
  urldate = {2022-01-06},
  archiveprefix = {arxiv}
}

@article{championBranchingTimeActive2021,
  title = {Branching {{Time Active Inference}}: Empirical Study and Complexity Class Analysis},
  shorttitle = {Branching {{Time Active Inference}}},
  author = {Champion, Th{\'e}ophile and Bowman, Howard and Grze{\'s}, Marek},
  year = {2021},
  month = nov,
  journal = {arXiv:2111.11276 [cs]},
  eprint = {2111.11276},
  primaryclass = {cs},
  urldate = {2022-01-03},
  archiveprefix = {arxiv}
}

@article{sajid2022active,
  title={Active inference, Bayesian optimal design, and expected utility},
  author={Sajid, Noor and Da Costa, Lancelot and Parr, Thomas and Friston, Karl},
  journal={The Drive for Knowledge: The Science of Human Information Seeking},
  pages={124},
  year={2022},
  publisher={Cambridge University Press}
}

@Article{Isomura2023,
  author   = {Isomura, Takuya and Kotani, Kiyoshi and Jimbo, Yasuhiko and Friston, Karl J.},
  journal  = {Nature Communications},
  title    = {Experimental validation of the free-energy principle with in vitro neural networks},
  year     = {2023},
  issn     = {2041-1723},
  number   = {1},
  pages    = {4547},
  volume   = {14},
  doi      = {10.1038/s41467-023-40141-z},
  refid    = {Isomura2023},
  url      = {https://doi.org/10.1038/s41467-023-40141-z},
}

@article{Paul2023,
title = {On efficient computation in active inference},
journal = {Expert Systems with Applications},
volume = {253},
pages = {124315},
year = {2024},
issn = {0957-4174},
url = {https://www.sciencedirect.com/science/article/pii/S0957417424011813},
author = {Aswin Paul and Noor Sajid and Lancelot {Da Costa} and Adeel Razi}
}

@Article{Paul2024,
AUTHOR = {Paul, Aswin and Isomura, Takuya and Razi, Adeel},
TITLE = {On Predictive Planning and Counterfactual Learning in Active Inference},
JOURNAL = {Entropy},
VOLUME = {26},
YEAR = {2024},
NUMBER = {6},
ARTICLE-NUMBER = {484},
URL = {https://www.mdpi.com/1099-4300/26/6/484},
PubMedID = {38920492},
ISSN = {1099-4300}
}

@Article{Kagan2022,
  author                 = {Kagan, Brett J. and Kitchen, Andy C. and Tran, Nhi T. and Habibollahi, Forough and Khajehnejad, Moein and Parker, Bradyn J. and Bhat, Anjali and Rollo, Ben and Razi, Adeel and Friston, Karl J.},
  journal                = {Neuron},
  title                  = {In vitro neurons learn and exhibit sentience when embodied in a simulated game-world.},
  year                   = {2022},
  month                  = {Dec},
  pages                  = {3952-3969.e8},
  volume                 = {110},
  electronic-issn        = {1097-4199},
  linking-issn           = {0896-6273},
  print-issn             = {0896-6273},
  address                = {United States},
  article-doi            = {10.1016/j.neuron.2022.09.001},
  article-pii            = {S0896-6273(22)00806-6},
  comment                = {Nature. 2022 Oct;610(7932):433. PMID: 36224373
Neuron. 2022 Dec 7;110(23):3855-3856. PMID: 36480938},
  completed              = {20221215},
  electronic-publication = {20221012},
  history                = {2022/10/13 18:42 [entrez]},
  issue                  = {23},
  language               = {eng},
  location-id            = {10.1016/j.neuron.2022.09.001 [doi]},
  nlm-unique-id          = {8809320},
  publication-status     = {ppublish},
  revised                = {20221220},
  source                 = {Neuron. 2022 Dec 7;110(23):3952-3969.e8. doi: 10.1016/j.neuron.2022.09.001. Epub 2022 Oct 12.},
  status                 = {PubMed-not-MEDLINE},
  subset                 = {IM},
  termowner              = {NOTNLM},
  title-abbreviation     = {Neuron},
}

@article{khajehnejad2026graph,
  title={Graph-based representation learning of neuronal dynamics and behavior},
  author={Khajehnejad, Moein and Habibollahi, Forough and Khajehnejad, Ahmad and French, Chris and Kagan, Brett J and Razi, Adeel},
  journal={Communications Biology},
  year={2026},
  publisher={Nature Publishing Group UK London}
}

\clearpage
\section*{Supplementary Information}
\setcounter{table}{0}
\renewcommand{\thetable}{S\arabic{table}}

\new{
This Supplementary Information collects the material supporting the analyses reported in the main text. The simulation protocol below records the software environment, the trial and episode structure, the seeding procedure and the definition of the two time windows, and the statistical framework that follows sets out the models, the resampling scheme and the correction applied to $p$-values. The tables are then grouped by what they report. Tables \ref{tab:within} to \ref{tab:absolute} give the within-group change over a trial and the level of performance reached, and illustrate why the two must be read together. Table \ref{tab:contrastsri} reports the agent--culture contrasts on relative improvement, together with the equivalence assessment. Tables \ref{tab:baselines} and \ref{tab:baseline-contrasts} place the four non-active-inference baselines of Sec.~\ref{sec:baseline-methods} beside the published agents and the two cultures. Tables \ref{tab:nte} and \ref{tab:ntepredict} report the entropy of the model parameters against the analytic maxima of Sec.~\ref{sec:nte}, and the across-seed association between the change in that entropy and the change in behaviour. Tables \ref{tab:hyperparams} to \ref{tab:noise} record the settings shared across the three schemes, the rate at which the learned transition model ceases to carry information as the rollout deepens, and performance under observation noise and unreliable action execution.
}

\subsection*{Simulation protocol}

\new{
This section records the detail needed to reproduce the simulations without recourse to the source repository.

\paragraph{Software environment}
All simulations and analyses were run under Python 3.10.11 with pinned dependencies: \texttt{numpy} 1.26.4, \texttt{pandas} 2.3.3, \texttt{scipy} 1.15.3, \texttt{statsmodels} 0.14.4, \texttt{matplotlib} 3.10.9 and \texttt{gym} 0.26.2. The environment specification is held in the project's \texttt{pyproject.toml} and lock file, so the versions above are reproduced exactly rather than resolved afresh.

\paragraph{Trial and episode structure}
A trial consists of $70$ Pong episodes, chosen to match the mean of $69.04 \pm 7.95$ episodes observed in the biological trials. We ran $100$ trials for each CFL-$T$ configuration and $99$ for each DP-$T$ configuration and for AIF-1. An episode ends when the environment signals termination, or when the cumulative reward within that episode exceeds $100$, whichever occurs first; the second condition bounds the cost of an episode in which the agent does not miss.

\paragraph{Seeding}
The seeding procedure is not uniform across the three schemes, and this matters for anyone attempting to reproduce a single trial. The DP-$T$ and AIF-1 runners seed both the Python and NumPy generators with the trial index at the start of each trial, and each trial is executed in its own process, so any individual trial can be reproduced in isolation. The CFL-$T$ runner instead seeds once at import and executes all $100$ trials within a single process, drawing from one continuous stream. A given CFL trial is therefore reproducible only by running the preceding trials in the same process and the same order.

\paragraph{Definition of the two time windows}
Each biological trial lasts twenty minutes and is divided into a first window of five minutes and a second of fifteen. The in-silico agents have no clock, so the division is made on elapsed simulation steps. We accumulate the number of steps taken up to the end of each episode, take one quarter of the total for the trial, and place in the first window every episode completed within that budget; the remaining episodes form the second window. The first window is then labelled as minutes $0$--$5$ and the second as minutes $6$--$20$. Every windowed quantity reported in this paper, for agents and for cultures alike, is computed against this division.
}

\subsection*{Statistical framework}

\new{
\paragraph{Models and uncertainty}
Within-group change from the first to the second window is estimated with a linear mixed-effects model, using the time window as a fixed effect and a random intercept on replicate. Between-group differences in relative improvement are estimated with a nonparametric bootstrap ($10^{5}$ resamples) that resamples whole replicates rather than individual sessions, so that repeated sessions from one culture cannot enter as independent evidence. Effect sizes are reported as Hedges' $g$. All $p$-values are corrected for multiple comparisons using the Benjamini--Hochberg procedure, applied across the full family of tests in each analysis, i.e. across all group $\times$ metric tests for the within-group models, and across all eighteen agent--culture contrasts for each between-group comparison. Every bootstrap is seeded deterministically, so the reported intervals are reproducible.

Throughout, we distinguish a contrast that is \textit{not distinguishable} from one that is \textit{equivalent}. The former means only that the difference cannot be resolved at this sample size, and is not evidence that the two groups match. To assess the latter we additionally test whether the whole confidence interval falls within a margin of $\pm 18$ percentage points of relative improvement, chosen as approximately the gap between the two culture types. We report this assessment in Supplementary Table \ref{tab:contrastsri} and note that no contrast attains equivalence under it.
}

\subsection*{Behaviour over the trial}

\begin{table}[ht]
\centering
\small
\caption{Change from the first (0--5 min) to the second (6--20 min) window, from a mixed-effects model with a random intercept on replicate. Each cell gives the estimate with its 95\% confidence interval and the corrected $p$-value.}
\label{tab:within}
\begin{tabular}{lllllll}
\toprule
Group & \multicolumn{2}{c}{$\Delta$ rally length} & \multicolumn{2}{c}{$\Delta$ \% long rallies} & \multicolumn{2}{c}{$\Delta$ \% aces} \\
\midrule
MCC & +0.147 [0.090, 0.204] & $<$0.001 & +3.421 [2.102, 4.740] & $<$0.001 & -1.580 [-3.948, 0.788] & 0.274 \\
HCC & +0.243 [0.190, 0.296] & $<$0.001 & +4.323 [3.426, 5.220] & $<$0.001 & -2.482 [-4.382, -0.583] & 0.026 \\
CFL-1 & +0.147 [-0.013, 0.307] & 0.139 & +0.577 [-1.287, 2.440] & 0.619 & -2.124 [-5.276, 1.029] & 0.274 \\
CFL-2 & +0.176 [0.063, 0.289] & 0.007 & +1.211 [-0.456, 2.877] & 0.255 & -0.648 [-3.859, 2.563] & 0.714 \\
CFL-3 & +0.450 [0.252, 0.648] & $<$0.001 & +2.788 [0.872, 4.704] & 0.012 & -2.326 [-5.400, 0.747] & 0.253 \\
CFL-4 & +1.158 [0.486, 1.830] & 0.002 & +5.674 [3.297, 8.051] & $<$0.001 & -5.314 [-8.342, -2.286] & 0.002 \\
AIF-1 & -0.016 [-0.088, 0.057] & 0.714 & +0.472 [-1.146, 2.090] & 0.624 & -1.055 [-4.021, 1.910] & 0.593 \\
DP-2 & +0.105 [-0.037, 0.248] & 0.255 & +1.063 [-0.835, 2.962] & 0.359 & -1.894 [-4.679, 0.890] & 0.274 \\
DP-5 & +0.304 [0.155, 0.453] & $<$0.001 & +0.331 [-1.740, 2.401] & 0.754 & -1.785 [-4.572, 1.001] & 0.288 \\
DP-10 & +0.354 [0.207, 0.501] & $<$0.001 & +1.961 [0.204, 3.717] & 0.059 & -1.301 [-4.161, 1.559] & 0.473 \\
DP-20 & +0.145 [0.025, 0.265] & 0.043 & +1.990 [0.290, 3.690] & 0.048 & +0.777 [-1.644, 3.198] & 0.619 \\
\bottomrule
\end{tabular}
\\[4pt]\footnotesize{Higher is better for rally length and long rallies; lower is better for aces. $p$-values are corrected across all 33 group $\times$ metric tests.}
\end{table}

\begin{table}[ht]
\centering
\caption{Average rally length reached in each window, the absolute gain between them, and the relative improvement. Relative improvement normalises by each session's own first-window mean, so it rewards a low starting point.}
\label{tab:level}
\begin{tabular}{lrrrr}
\toprule
Group & Window 1 & Window 2 & Absolute gain & Relative improvement (\%) \\
\midrule
CFL-4 & 1.325 & 2.482 & +1.157 & +106.2 \\
CFL-3 & 0.988 & 1.439 & +0.450 & +75.6 \\
DP-5 & 0.885 & 1.189 & +0.304 & +70.9 \\
DP-10 & 0.825 & 1.179 & +0.354 & +66.9 \\
CFL-1 & 0.873 & 1.019 & +0.147 & +47.2 \\
CFL-2 & 0.835 & 1.011 & +0.176 & +40.7 \\
DP-20 & 0.835 & 0.980 & +0.145 & +40.0 \\
DP-2 & 0.852 & 0.957 & +0.105 & +49.0 \\
HCC & 0.637 & 0.880 & +0.243 & +57.1 \\
MCC & 0.700 & 0.847 & +0.147 & +39.3 \\
AIF-1 & 0.817 & 0.801 & -0.016 & +15.8 \\
\bottomrule
\end{tabular}
\\[4pt]\footnotesize{AIF-1 illustrates the limitation of the relative measure: its absolute change is negative ($-0.016$) while its mean relative improvement is positive ($+15.8\%$), because the mean of per-session ratios is not the ratio of means and is skewed by sessions with small first-window denominators.}
\end{table}

\begin{table}[ht]
\centering
\small
\caption{Absolute performance in the second window (6--20 min), with 95\% intervals from a replicate-level bootstrap.}
\label{tab:absolute}
\begin{tabular}{llll}
\toprule
Group & Rally length & \% long rallies & \% aces \\
\midrule
MCC & 0.85 [0.81, 0.91] & 9.23 [8.69, 10.49] & 54.26 [52.35, 55.55] \\
HCC & 0.88 [0.82, 0.95] & 8.06 [7.17, 8.95] & 53.94 [52.54, 55.07] \\
CFL-1 & 1.02 [0.90, 1.18] & 11.19 [10.13, 12.28] & 56.75 [54.82, 58.65] \\
CFL-2 & 1.01 [0.92, 1.11] & 11.10 [9.95, 12.29] & 57.09 [54.66, 59.51] \\
CFL-3 & 1.44 [1.26, 1.64] & 14.10 [12.64, 15.65] & 54.19 [51.80, 56.56] \\
CFL-4 & 2.47 [1.80, 3.59] & 19.05 [17.08, 21.36] & 47.03 [44.75, 49.25] \\
AIF-1 & 0.80 [0.77, 0.84] & 8.47 [7.66, 9.32] & 54.27 [52.79, 55.74] \\
DP-2 & 0.96 [0.86, 1.07] & 11.16 [10.27, 12.08] & 58.58 [56.68, 60.47] \\
DP-5 & 1.19 [1.06, 1.33] & 11.15 [10.29, 12.03] & 57.85 [56.27, 59.43] \\
DP-10 & 1.18 [1.05, 1.32] & 11.69 [10.76, 12.63] & 57.32 [55.77, 58.91] \\
DP-20 & 0.98 [0.89, 1.08] & 11.70 [10.74, 12.67] & 59.70 [57.70, 61.69] \\
\bottomrule
\end{tabular}
\\[4pt]\footnotesize{Higher is better for rally length and long rallies; lower is better for aces. On aces, CFL-1 and CFL-2 are worse than both cultures.}
\end{table}

\newpage
\subsection*{Agent-culture contrasts}

\begin{table}[ht]
\centering
\small
\caption{Relative improvement of each in-silico agent against the two cultures, reported for continuity with the original DishBrain analysis. Outcomes are three-way: ``different'' where the contrast is significant after correction, ``equivalent'' where the interval falls entirely within a margin of $\pm18$ percentage points (approximately the MCC--HCC gap), and ``inconclusive'' otherwise.}
\label{tab:contrastsri}
\begin{tabular}{llrrrrl}
\toprule
Agent & Culture & $\Delta$RI (\%) & 95\% CI & Hedges' $g$ & $p_{\mathrm{adj}}$ & Outcome \\
\midrule
CFL-1 & MCC & +4.0 & [-31.6, 35.6] & +0.08 & 0.875 & inconclusive \\
CFL-1 & HCC & -10.1 & [-42.1, 21.2] & -0.11 & 0.725 & inconclusive \\
CFL-2 & MCC & -2.5 & [-35.5, 24.9] & +0.02 & 0.952 & inconclusive \\
CFL-2 & HCC & -16.6 & [-46.2, 10.4] & -0.21 & 0.481 & inconclusive \\
CFL-3 & MCC & +32.4 & [-4.0, 65.2] & +0.36 & 0.298 & inconclusive \\
CFL-3 & HCC & +18.3 & [-15.1, 50.7] & +0.19 & 0.482 & inconclusive \\
CFL-4 & MCC & +63.0 & [23.3, 100.4] & +0.58 & 0.007 & agent higher \\
CFL-4 & HCC & +48.9 & [11.8, 85.6] & +0.45 & 0.054 & inconclusive \\
AIF-1 & MCC & -27.4 & [-57.7, -4.7] & -0.36 & 0.054 & inconclusive \\
AIF-1 & HCC & -41.5 & [-68.5, -18.2] & -0.65 & 0.007 & agent lower \\
DP-2 & MCC & +5.8 & [-29.0, 35.8] & +0.11 & 0.828 & inconclusive \\
DP-2 & HCC & -8.3 & [-39.8, 21.4] & -0.09 & 0.763 & inconclusive \\
DP-5 & MCC & +27.7 & [-12.2, 67.2] & +0.27 & 0.469 & inconclusive \\
DP-5 & HCC & +13.7 & [-23.3, 52.5] & +0.12 & 0.725 & inconclusive \\
DP-10 & MCC & +23.7 & [-10.9, 53.5] & +0.31 & 0.469 & inconclusive \\
DP-10 & HCC & +9.6 & [-21.6, 39.0] & +0.12 & 0.725 & inconclusive \\
DP-20 & MCC & -3.3 & [-36.0, 23.4] & +0.01 & 0.952 & inconclusive \\
DP-20 & HCC & -17.4 & [-46.8, 9.4] & -0.23 & 0.476 & inconclusive \\
\bottomrule
\end{tabular}
\\[4pt]\footnotesize{No contrast reaches equivalence. The measure also fails to separate the two cultures from each other (HCC $-$ MCC $= +14.1$ points, 95\% CI [-20.9, 45.9]), which sets the resolution floor for any comparison made with it.}
\end{table}

\newpage
\subsection*{Baselines outside active inference}

\begin{table}[ht]
\centering
\small
\caption{Non-active-inference baselines beside the published agents and the cultures. Absolute rally length is the second-window mean; the change across the trial is the paired within-session difference. Intervals are replicate-level bootstraps.}
\label{tab:baselines}
\begin{tabular}{lllll}
\toprule
Class & Group & Rally length (6--20 min) & Change across trial & $p_{\mathrm{adj}}$ \\
\midrule
In vitro & MCC & 0.85 [0.81, 0.91] & +0.15 [0.08, 0.29] & $<$0.001 \\
In vitro & HCC & 0.88 [0.82, 0.95] & +0.24 [0.12, 0.36] & $<$0.001 \\
Active inference & AIF-1 & 0.80 [0.77, 0.84] & -0.02 [-0.09, 0.06] & 0.682 \\
Active inference & CFL-1 & 1.02 [0.90, 1.18] & +0.15 [0.00, 0.32] & 0.093 \\
Active inference & CFL-2 & 1.01 [0.92, 1.11] & +0.18 [0.07, 0.29] & 0.005 \\
Active inference & CFL-3 & 1.44 [1.26, 1.64] & +0.45 [0.26, 0.65] & $<$0.001 \\
Active inference & CFL-4 & 2.47 [1.80, 3.59] & +1.16 [0.64, 1.93] & 0.002 \\
Active inference & DP-2 & 0.96 [0.86, 1.07] & +0.11 [-0.03, 0.25] & 0.175 \\
Active inference & DP-5 & 1.19 [1.06, 1.33] & +0.30 [0.16, 0.46] & $<$0.001 \\
Active inference & DP-10 & 1.18 [1.05, 1.32] & +0.35 [0.21, 0.51] & $<$0.001 \\
Active inference & DP-20 & 0.98 [0.89, 1.08] & +0.14 [0.03, 0.27] & 0.030 \\
Baseline & Random & 0.80 [0.77, 0.83] & +0.02 [-0.04, 0.08] & 0.640 \\
Baseline & Heuristic & 12.07 [11.28, 12.89] & +1.67 [-0.13, 3.34] & 0.087 \\
Baseline & DynaQ & 1.17 [1.10, 1.24] & +0.28 [0.20, 0.37] & $<$0.001 \\
Baseline & DynaQ-replay & 1.04 [1.00, 1.09] & +0.17 [0.09, 0.24] & $<$0.001 \\
\bottomrule
\end{tabular}
\\[4pt]\footnotesize{The random controller cannot learn, and its change across the trial is accordingly indistinguishable from zero, which calibrates the improvement metric. The reactive heuristic does not learn either, but is well matched to the task.}
\end{table}

\begin{table}[ht]
\centering
\small
\caption{Baseline contrasts on second-window rally length. Benjamini-Hochberg correction is applied within this family of twelve comparisons, which is separate from the agent-versus-culture family of the main analysis.}
\label{tab:baseline-contrasts}
\begin{tabular}{llrrrrl}
\toprule
A & B & $\Delta$ rally & 95\% CI & Hedges' $g$ & $p_{\mathrm{adj}}$ & Outcome \\
\midrule
DynaQ & CFL-1 & +0.15 & [-0.03, 0.29] & +0.26 & 0.108 & comparable \\
DynaQ & CFL-2 & +0.16 & [0.04, 0.27] & +0.36 & 0.018 & A higher \\
DynaQ & CFL-3 & -0.27 & [-0.48, -0.08] & -0.37 & 0.009 & B higher \\
DynaQ & CFL-4 & -1.30 & [-2.43, -0.62] & -0.37 & $<$0.001 & B higher \\
DynaQ & DP-5 & -0.02 & [-0.18, 0.12] & -0.04 & 0.861 & comparable \\
DynaQ & AIF-1 & +0.37 & [0.29, 0.44] & +1.33 & $<$0.001 & A higher \\
Random & AIF-1 & +0.00 & [-0.05, 0.05] & +0.01 & 0.952 & comparable \\
Random & MCC & -0.04 & [-0.11, 0.00] & -0.26 & 0.080 & comparable \\
Random & HCC & -0.08 & [-0.15, -0.01] & -0.43 & 0.033 & B higher \\
Heuristic & CFL-4 & +9.60 & [8.24, 10.75] & +2.10 & $<$0.001 & A higher \\
DynaQ & MCC & +0.32 & [0.23, 0.40] & +1.16 & $<$0.001 & A higher \\
DynaQ & HCC & +0.29 & [0.19, 0.38] & +1.07 & $<$0.001 & A higher \\
\bottomrule
\end{tabular}
\\[4pt]\footnotesize{``Comparable'' denotes an absence of a detectable difference, not evidence of equivalence.}
\end{table}

\newpage
\subsection*{Parameter entropy}

\begin{table}[ht]
\centering
\small
\caption{Normalised total entropy against the analytic maximum, by time window. NTE is one when the parameter is uniform and approaches zero as it becomes fully determined, so values are comparable across agents and parameters.}
\label{tab:nte}
\begin{tabular}{lllrrrr}
\toprule
Agent & Parameter & $\mathrm{TE}_{\max}$ & NTE 0--5 & NTE 6--20 & $\Delta$ & Fall (\% of max) \\
\midrule
CFL-1 & $\mathbb{CL}$ & 2671.83 & 0.9987 & 0.9956 & -0.0031 & 1.2 \\
CFL-2 & $\mathbb{CL}$ & 2671.83 & 0.9976 & 0.9922 & -0.0054 & 2.0 \\
CFL-3 & $\mathbb{CL}$ & 2671.83 & 0.9960 & 0.9884 & -0.0076 & 2.9 \\
CFL-4 & $\mathbb{CL}$ & 2671.83 & 0.9936 & 0.9814 & -0.0122 & 4.3 \\
AIF-1 & $\mathbb{B}$ (ball-x) & 414.68 & 0.4148 & 0.3616 & -0.0532 & 83.8 \\
DP-2 & $\mathbb{B}$ (ball-x) & 414.68 & 0.2758 & 0.2177 & -0.0580 & 71.3 \\
DP-5 & $\mathbb{B}$ (ball-x) & 414.68 & 0.3566 & 0.3159 & -0.0407 & 68.3 \\
DP-10 & $\mathbb{B}$ (ball-x) & 414.68 & 0.3584 & 0.3239 & -0.0345 & 68.0 \\
DP-20 & $\mathbb{B}$ (ball-x) & 414.68 & 0.3570 & 0.3187 & -0.0383 & 68.0 \\
\bottomrule
\end{tabular}
\\[4pt]\footnotesize{The final column is the fall from the highest to the lowest value attained over a trial, expressed as a percentage of the analytic maximum. The planning schemes move their transition parameters by an order of magnitude more than the counterfactual scheme moves its mapping, and the size of that movement does not order the agents by performance.}
\end{table}

\begin{table}[ht]
\centering
\small
\caption{Across-seed association between the change in parameter entropy over a trial and the change in average rally length. A negative coefficient would indicate that seeds whose parameters became more informed improved more.}
\label{tab:ntepredict}
\begin{tabular}{lrrrrr}
\toprule
Agent & Seeds & Pearson $r$ & $p_{\mathrm{adj}}$ & Spearman $\rho$ & $p_{\mathrm{adj}}$ \\
\midrule
CFL-1 & 100 & -0.205 & 0.238 & -0.122 & 0.941 \\
CFL-2 & 100 & -0.168 & 0.284 & -0.008 & 0.941 \\
CFL-3 & 100 & -0.054 & 0.671 & +0.024 & 0.941 \\
CFL-4 & 100 & -0.119 & 0.534 & +0.040 & 0.941 \\
AIF-1 & 99 & +0.062 & 0.671 & +0.028 & 0.941 \\
DP-2 & 99 & +0.070 & 0.671 & +0.015 & 0.941 \\
DP-5 & 99 & +0.042 & 0.681 & +0.021 & 0.941 \\
DP-10 & 99 & +0.195 & 0.238 & +0.203 & 0.395 \\
DP-20 & 99 & -0.097 & 0.610 & -0.094 & 0.941 \\
\bottomrule
\end{tabular}
\\[4pt]\footnotesize{No association survives correction for either coefficient. Parameter entropy change therefore does not predict behavioural improvement within any agent.}
\end{table}

\subsection*{Model settings and robustness}

\begin{table}[ht]
\centering
\small
\caption{Hyperparameters of the three decision-making schemes, read directly from the simulation code. All settings that the schemes share are identical; the quantities that differ are the ones that define the schemes, together with the prior specification of AIF-1.}
\label{tab:hyperparams}
\begin{tabular}{lllll}
\toprule
Setting & CFL-$T$ & DP-$T$ & AIF-1 & Status \\
\midrule
Episodes per trial & 70 & 70 & 70 & shared \\
Action precision & 1024 & 1024 & 1024 & shared \\
Transition learning rate & $10^{16}$ & $10^{16}$ & $10^{16}$ & shared \\
Numerical floor & $10^{-16}$ & $10^{-16}$ & $10^{-16}$ & shared \\
Prior preferences $\mathbb{C}$ & uniform & uniform & random categorical & differs \\
Initial state prior $\mathbb{D}$ & uniform & uniform & random categorical & differs \\
Horizon & memory, $T \in \{1,2,3,4\}$ & planning, $T \in \{2,5,10,20\}$ & planning, $T = 1$ & differs \\
\bottomrule
\end{tabular}
\\[4pt]\footnotesize{The horizon is the quantity under study and differs by construction. The prior specification of AIF-1 is the one unintended asymmetry; its effect is measured in Sec.~\ref{sec:priorcheck} and found not to account for the performance of that agent.}
\end{table}

\begin{table}[ht]
\centering
\small
\caption{Predictive information of the learned transition model as a function of rollout depth. Entries are the entropy of the $h$-step-ahead prediction, normalised by the entropy of the uniform distribution over the same support. The value of one means the rollout carries no information about the future state. Here, normalised entropy is computed from the transition parameters of the DP-5 agent on 3 seeds, averaged over the three actions and over all starting states. The environment itself is deterministic; what decays with depth is the information carried by the agent's \textit{estimate} of it.}
\label{tab:planningdepth}
\begin{tabular}{rrr}
\toprule
Depth $h$ & Normalised entropy & SD \\
\midrule
1 & 0.324 & 0.054 \\
2 & 0.513 & 0.066 \\
3 & 0.605 & 0.071 \\
4 & 0.666 & 0.075 \\
5 & 0.707 & 0.078 \\
10 & 0.810 & 0.088 \\
15 & 0.850 & 0.096 \\
20 & 0.868 & 0.102 \\
\bottomrule
\end{tabular}
\end{table}

\begin{table}[ht]
\centering
\small
\caption{Performance under observation noise and unreliable action execution. The first row of each block is the control condition, in which both manipulations are switched off and the published run is recovered.}
\label{tab:noise}
\begin{tabular}{llrrrrrr}
\toprule
Agent & Condition & Seeds & 0--5 min & 6--20 min & Median & $\Delta$ & $p$ \\
\midrule
CFL-4 & control & 100 & 1.08 & 2.39 & 1.83 & +1.31 & $<$0.001 \\
CFL-4 & obs. $\varepsilon = 0.05$ & 100 & 2.40 & 6.29 & 2.81 & +3.89 & $<$0.001 \\
CFL-4 & obs. $\varepsilon = 0.15$ & 100 & 1.94 & 4.75 & 3.27 & +2.81 & $<$0.001 \\
CFL-4 & act. $\eta = 0.10$ & 100 & 1.01 & 1.46 & 1.24 & +0.44 & $<$0.001 \\
DP-5 & control & 6 & 0.91 & 1.30 & 1.21 & +0.38 & 0.273 \\
DP-5 & obs. $\varepsilon = 0.05$ & 6 & 0.85 & 1.12 & 1.10 & +0.27 & 0.117 \\
DP-5 & obs. $\varepsilon = 0.15$ & 6 & 0.85 & 0.90 & 0.94 & +0.04 & 0.812 \\
DP-5 & act. $\eta = 0.10$ & 6 & 1.23 & 0.73 & 0.73 & -0.49 & 0.018 \\
\bottomrule
\end{tabular}
\\[4pt]\footnotesize{The median is given because the distribution of rally length over seeds is heavy tailed for CFL-4, so the mean alone would overstate the typical run. The DP-5 rows rest on six seeds and are indicative only; no difference between its observation-noise conditions is distinguishable from zero.}
\end{table}

\end{document}